\documentclass[11pt]{article}
\PassOptionsToPackage{table}{xcolor}
\PassOptionsToPackage{hyperfootnotes=false}{hyperref}
\ifdefined\pdfminorversion\pdfminorversion=7\fi

\usepackage[preprint]{acl}

\usepackage{times}
\usepackage{latexsym}

\usepackage{algorithm}
\usepackage{algorithmic}
\usepackage{amsmath}
\usepackage{booktabs}
\usepackage{multirow}
\usepackage{graphicx}
\usepackage{amssymb}
\usepackage{tabularx}
\usepackage{array}
\usepackage{enumitem}
\newlist{promptenum}{enumerate}{1}
\setlist[promptenum]{label=\arabic*.,leftmargin=*,topsep=3pt,itemsep=2pt,parsep=0pt}

\usepackage{listings}

\usepackage{tcolorbox}
\tcbuselibrary{skins,breakable}

\definecolor{promptblue}{RGB}{82,121,163}      
\definecolor{promptframe}{RGB}{120,150,180}    
\definecolor{promptbg}{RGB}{243,247,252}       

\newtcolorbox{promptbox}[1]{
    enhanced,
    breakable,
    colback=promptbg,
    colframe=promptframe,
    boxrule=0.7pt,
    arc=1.5pt,
    colbacktitle=promptblue,
    coltitle=white,
    title={#1},
    fonttitle=\footnotesize\bfseries,
    toptitle=3pt,
    bottomtitle=3pt,
    lefttitle=1pt,
    left=5pt,
    right=5pt,
    top=4pt,
    bottom=4pt,
    before skip=5pt,
    after skip=5pt,
    before upper={\small\raggedright},
    parbox=false
}
\usepackage[T1]{fontenc}
\usepackage[utf8]{inputenc}

\usepackage{microtype}
\microtypesetup{stretch=30,shrink=20}
\usepackage{hyphenat}
\AddToHook{env/thebibliography/begin}{\raggedright}

\usepackage{inconsolata}

\usepackage{graphicx}

\title{CALM: Conflict\hyp{}Aware Lifecycle Memory Governance for Long\hyp{}Horizon Language Agents}

\author{
  \textbf{Chang Li\textsuperscript{1}},
  \textbf{Xianhao Ou\textsuperscript{1}},
  \textbf{Xinyan Ma\textsuperscript{1}},
  \textbf{Hao Li\textsuperscript{1}},
  \textbf{Haoyuan Xu\textsuperscript{1}} \\
  \textbf{Jiafeng Liang\textsuperscript{1}},
  \textbf{Xiaohan Wang\textsuperscript{2}},
  \textbf{Jiajun Chai\textsuperscript{2}},
  \textbf{Guojun Yin\textsuperscript{2}},
    \textbf{Ming Liu\textsuperscript{1}}\thanks{Corresponding author.} \\[0.5em]
  \textsuperscript{1}Harbin Institute of Technology
  \quad
  \textsuperscript{2}Meituan
}

\begin{document}
\fussy
\tolerance=1000
\emergencystretch=1em
\raggedbottom
\maketitle
\begin{abstract}

A memory can be outdated for present\hyp{}day reasoning yet indispensable for reconstructing the past. Relevance\hyp{}based retrieval overlooks this tension: semantic relevance, evidence\hyp{}grounded validity, and query\hyp{}specific eligibility are distinct. We propose CALM, a conflict\hyp{}aware lifecycle memory governance framework that decouples persistent memory preservation from query\hyp{}time admission. CALM converts heterogeneous records into evidence\hyp{}grounded memory cells, updates validity and lifecycle states using support, conflict, and dependency signals, and admits query\hyp{}appropriate evidence through training\hyp{}free or learned gating informed by temporal applicability. Across LongMemEval\hyp{}S and LoCoMo with Qwen and GPT backbones, CALM outperforms state\hyp{}of\hyp{}the\hyp{}art baselines, achieving the best overall accuracy. Compared with simple filtering and reranking, query\hyp{}conditioned admission further improves evidence selection. Conflict\hyp{}blocking and robustness analyses show CALM suppresses erroneous retrieved memories while retaining useful evidence. Historical\hyp{}reconstruction experiments demonstrate that preserving superseded records supports accurate earlier\hyp{}state reasoning without compromising current\hyp{}state answers.

\end{abstract}

\section{Introduction}

For long\hyp{}horizon language agents, reliable memory requires deciding not only what to retain, but also what to use. As users relocate, revise preferences, or change commitments, interaction histories accumulate records that remain informative about the past but may no longer describe the present. Existing memory\hyp{}augmented systems have advanced the storage, organization, and retrieval of such histories~\citep{Intro_memgpt,Intro_membank,Intro_hipporag,Intro_zep}. Yet semantic relevance does not imply query eligibility: a retrieved memory can be topically related to a question without being reliable or applicable to the state being asked about~\citep{tma,agent_memory_survey}. This mismatch can turn successful retrieval into misleading reasoning.
\begin{figure}[t]
    \centering
    \includegraphics[width=\linewidth]{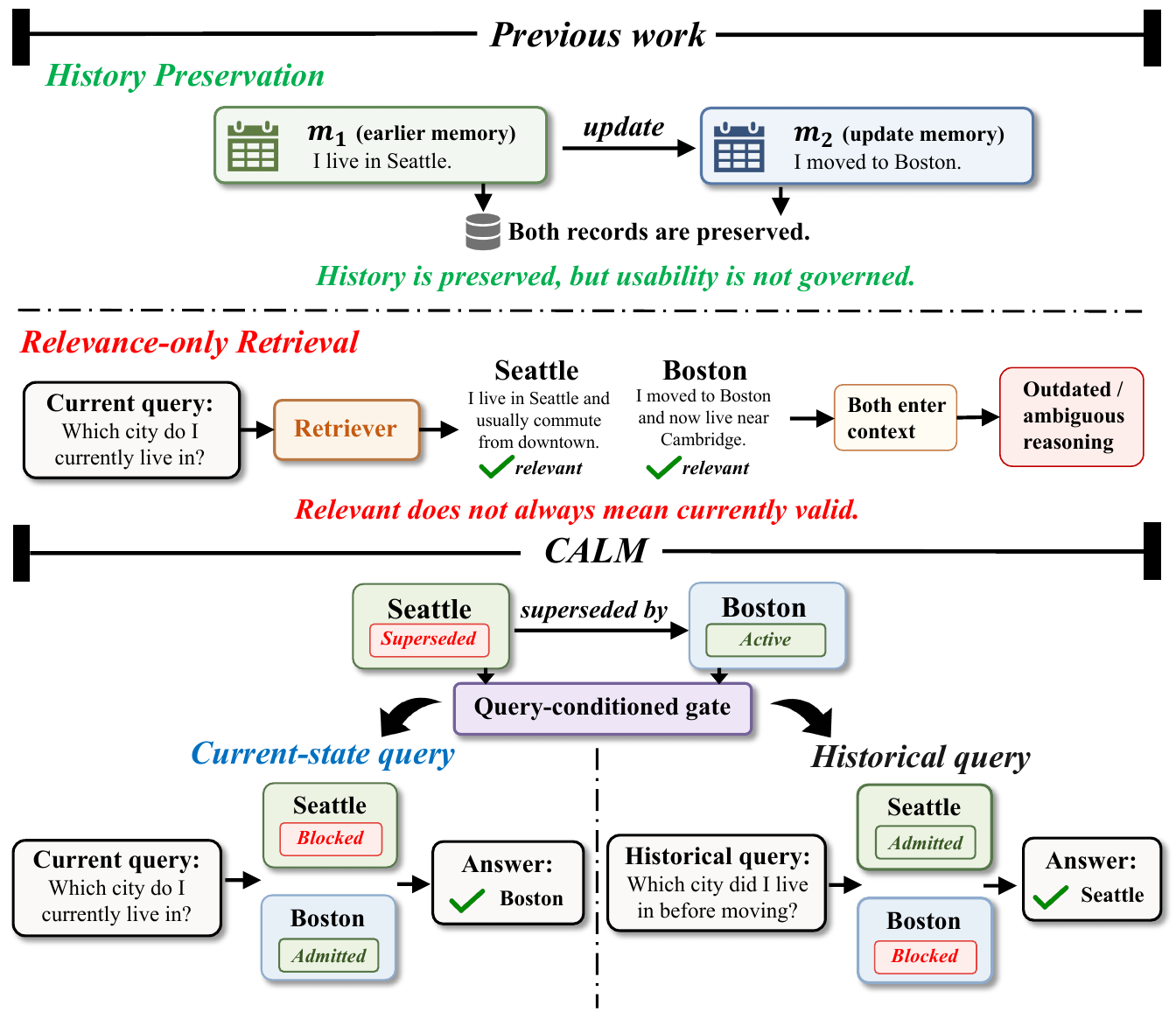}
    \caption{Schematic comparison of relevance\hyp{}only retrieval and CALM. CALM retains earlier records for historical queries while excluding incompatible evidence from current\hyp{}state reasoning.}
    \label{figure_teaser}
\end{figure}

Consider a user who moves from Seattle to Boston. The earlier residence should not determine the answer to a current\hyp{}location query, but it remains necessary for a question about where the user previously lived. As illustrated in Figure~\ref{figure_teaser}, deleting the earlier record loses historical evidence, whereas admitting both records indiscriminately can obscure the current state. This example exposes three distinct properties of memory: \emph{semantic relevance} measures relatedness to a query; \emph{query\hyp{}independent validity} reflects the memory's persistent, evidence\hyp{}governed status; and \emph{query\hyp{}conditioned eligibility} determines whether it should enter the working context given the query's temporal target, applicability scope, and intent. Even a superseded memory may be eligible for historical reconstruction, while an otherwise well\hyp{}supported memory may still be ineligible when its time or scope does not match the query.

Graph\hyp{}based and temporal memory systems can represent evolving memory relationships, while recent work further highlights the challenges posed by conflicting records~\citep{Intro_hipporag,Intro_amem,Intro_zep,Intro_memconflict}. However, recording an update or conflict does not, by itself, resolve which retrieved records should be admitted for a particular question. A memory\hyp{}use policy must account for both persistent evidence and the conditions imposed by the query. This calls for separating lifecycle governance from query\hyp{}conditioned admission, rather than relying on memory organization or relevance\hyp{}based ranking alone.

We introduce Conflict\hyp{}Aware Lifecycle Memory Governance (CALM), a two\hyp{}level framework separating persistent lifecycle governance from query\hyp{}conditioned admission. Between the memory backend and working context, CALM adapts heterogeneous records into evidence\hyp{}grounded cells while preserving provenance and temporal information. It updates stored validity and lifecycle states using deterministic support, conflict, and dependency rules, with temporal cues guiding applicability rather than automatic decay. At query time, a shared hard eligibility filter precedes either a training\hyp{}free policy (CALM\hyp{}ZS) or a learnable gate (CALM). Both select evidence using retrieval relevance, stored validity, query--memory compatibility, and query\hyp{}conditioned risk. CALM thus retains superseded records for historical queries while excluding them from current\hyp{}state reasoning, without modifying the underlying retriever or answer model.

We evaluate CALM on LongMemEval\hyp{}S~\citep{longmemeval} and LoCoMo~\citep{locomo} across Qwen and GPT backbones. CALM achieves the best or tied\hyp{}best overall accuracy across evaluated settings, reaching 79.60\% on LongMemEval\hyp{}S with Qwen3-30B\hyp{}A3B\hyp{}Instruct-2507. Controlled comparisons with simple filtering and reranking demonstrate the benefits of combining lifecycle governance with query\hyp{}conditioned admission. Conflict\hyp{}blocking and robustness analyses show that CALM suppresses misleading evidence while preserving useful memories; historical reconstruction confirms that superseded information remains accessible without compromising current\hyp{}state reasoning. Component ablations and efficiency analyses further highlight the complementary roles of persistent governance and query\hyp{}aware selection. 

Our contributions are as follows:
\begin{itemize}[leftmargin=*, labelindent=0pt, labelsep=0.5em]
    \item We formulate long\hyp{}horizon memory use as a query\hyp{}conditioned eligibility problem, distinguishing semantic relevance, query\hyp{}independent validity, and query\hyp{}specific admission while preserving historical evidence.

    \item We propose CALM, which couples evidence\hyp{}grounded lifecycle governance with compatibility and risk\hyp{}aware admission, and instantiate both training\hyp{}free and learned gating policies.

    \item We validate CALM across two benchmarks and three answer\hyp{}model backbones through comparisons with existing memory systems and simple alternatives, alongside robustness, historical\hyp{}reconstruction, and efficiency analyses.
\end{itemize}

\section{Related Work}
\subsection{Long\hyp{}Term Memory for LLM Agents} 
Long\hyp{}term memory enables LLM agents to retain interaction histories, personalize responses, and support planning and reasoning across sessions. 
Early agents retain observations, feedback, and acquired skills to support cross\hyp{}session planning~\citep{Intro_generative,reflexion,voyager}.
MemoryBank~\citep{Intro_membank} maintains user information, while MemGPT~\citep{Intro_memgpt} pages context across hierarchical memory tiers; external\hyp{}memory architectures further scale recall~\citep{rel_aug}.
However, persistent retrieval alone does not ensure that past evidence is valid or applicable to the current query.

\subsection{Structured Memory Systems}
Recent systems organize memories through linked notes, graphs, and hierarchies~\citep{arigraph,hmem,nemori}.
HippoRAG~\citep{Intro_hipporag} combines knowledge graphs with Personalized PageRank for associative multi\hyp{}hop retrieval, while A\hyp{}Mem~\citep{Intro_amem} dynamically links and evolves memory notes.
Mem0~\citep{mem0} consolidates conversational facts through add, update, and delete operations.
MemoryOS, MemOS, and EverMemOS~\citep{memoryos,memos,evermemos} explore hierarchical management and memory lifecycles.
While HippoRAG emphasizes associative retrieval, A\hyp{}Mem focuses on adaptive memory organization, and Mem0 handles incremental memory updates, CALM focuses on post\hyp{}retrieval eligibility control.

\subsection{Temporal Consistency and Memory Governance}
\label{sec:temporal_governance}

Prior work studies memory staleness, conflicts, and interference ~\citep{stale,Intro_memconflict,tma,govern,agent_age,survey_long,from_untrust}. Zep~\citep{Intro_zep} tracks superseded facts with bi\hyp{}temporal graphs, while Mem0~\citep{mem0} resolves conflicts through memory updates.
A\hyp{}TMA~\citep{tma} coordinates memory\hyp{}bank maintenance, state\hyp{}aware evidence expansion, and answer\hyp{}time resolution. In contrast, CALM targets \emph{post\hyp{}retrieval admission}: it maintains evidence\hyp{}grounded, query\hyp{}independent validity and lifecycle states, including dependency invalidation, then applies hard eligibility constraints and training\hyp{}free
or learned gates to fixed retrieved candidates. Unlike A\hyp{}TMA's evidence expansion and answer\hyp{}side state resolution, CALM explicitly governs which existing memories enter the working context without modifying upstream retrieval or the answer model.
\section{CALM Architecture}

\paragraph{Problem Setting.}
A long\hyp{}horizon language agent accumulates an interaction history
\[
\mathcal{H}_t=\{(x_\ell,y_\ell,\tau_\ell)\}_{\ell=1}^{t},
\]
where $x_\ell$, $y_\ell$, and $\tau_\ell$ denote the user input, agent response, and timestamp at step $\ell$, respectively. An existing memory backend transforms this history into persistent records $\widetilde{\mathcal{M}}_t$, which CALM adapts into canonical memory cells $\mathcal{M}_t$. Given a query $q$, an upstream retriever provides candidate memories, represented after adaptation as $\mathcal{R}_q\subseteq\mathcal{M}_t$. For each retrieved memory $m_i^t\in\mathcal{R}_q$, we distinguish three properties: semantic relevance $\zeta_{q,i}^t\in[0,1]$ measures its relatedness to the query; query\hyp{}independent validity $V_i^t\in[0,1]$ reflects its reliability based on accumulated evidence; and query\hyp{}conditioned eligibility $g_{q,i}^t\in\{0,1\}$ determines whether it should be admitted under the query's temporal target, applicability scope, and intent. These properties are distinct: a relevant memory may be superseded or contradicted, while a superseded memory may remain eligible for historical queries. Accordingly, the working context is defined as $\mathcal{C}_q=\{m_i^t\in\mathcal{R}_q\mid g_{q,i}^t=1\}$, from which the agent generates $\hat{a}=f_{\mathrm{LLM}}(q,\mathcal{C}_q)$.

\begin{figure*}[t]
    \centering
    \includegraphics[width=\linewidth]{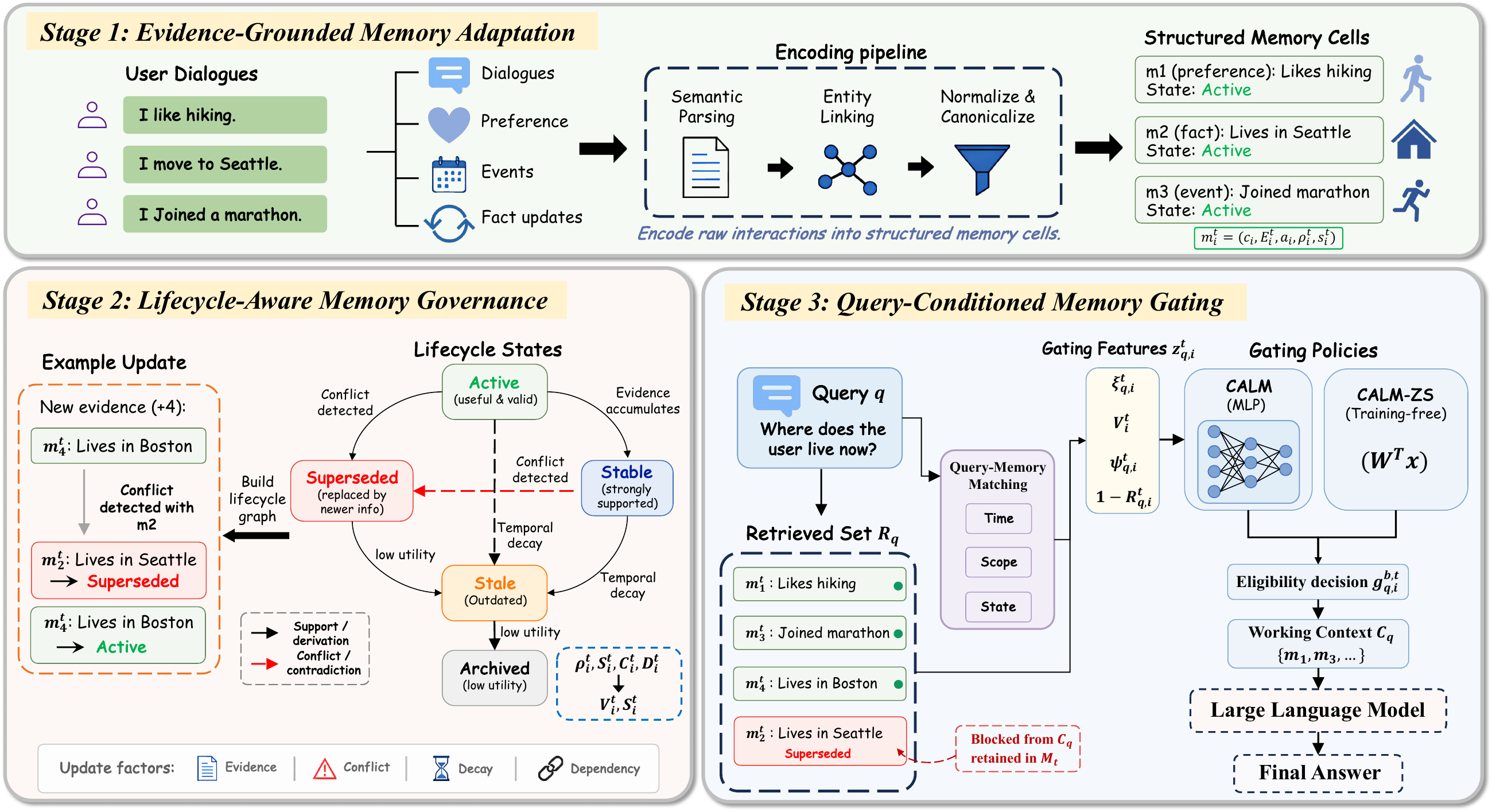}
    \caption{Overview of CALM. (a) Evidence\hyp{}Grounded Memory Adaptation converts heterogeneous backend records into canonical memory cells. (b) Lifecycle\hyp{}Aware Memory Governance updates their validity and lifecycle states using support, conflict, dependency, and temporal signals. (c) Query\hyp{}Conditioned Memory Gating admits eligible memories via \textsc{CALM\hyp{}ZS} or \textsc{CALM}, while retaining non\hyp{}admitted memories in persistent storage for traceability and future compatible queries.
    }
    \label{fig111}
\end{figure*}

\paragraph{Framework Overview.}
Inspired by the adaptive lifecycle of human memory~\citep{method_human_memory,Intro_fearmem,Intro_reconsolidation}, we propose CALM, a conflict\hyp{}aware lifecycle memory governance framework that separates persistent memory preservation from query\hyp{}time utilization. As illustrated in Figure~\ref{fig111}, CALM comprises three stages: \emph{Evidence\hyp{}Grounded Memory Adaptation} converts heterogeneous backend records into canonical memory cells; \emph{Lifecycle\hyp{}Aware Memory Governance} maintains evidence\hyp{}grounded validity and lifecycle states through support, conflict, dependency, and temporal signals; and \emph{Query\hyp{}Conditioned Memory Gating} selects eligible memories using either the training\hyp{}free \textsc{CALM\hyp{}ZS} rule or the learnable \textsc{CALM} policy. Rather than replacing the underlying storage or retrieval architecture, CALM separates persistent lifecycle governance from query\hyp{}local admission decisions, allowing superseded memories to remain available for historical reconstruction without automatically entering current\hyp{}state reasoning.

\subsection{Stage I: Evidence\hyp{}Grounded Memory Adaptation}
\label{sec:memory-adaptation}

Memory backends represent interaction histories as dialogue fragments, extracted facts, summaries, or structured records, which differ in granularity and metadata. To enable consistent memory governance, CALM converts these heterogeneous backend records into canonical memory cells while preserving their original evidence provenance:
\begin{equation}
\label{eq:memory-cell}
\begin{aligned}
m_i^t &= (c_i,\mathcal{E}_i^t,a_i,\rho_i^t,s_i^t),\\
a_i &= (\kappa_i,\tau_i,\pi_i,U_i,\Omega_i).
\end{aligned}
\end{equation}
Here, $c_i$ denotes the canonical content of a memory cell, which may correspond to an extracted proposition or a retained source span depending on the available structure of the backend record.
The attribute tuple $a_i$ specifies semantic type $\kappa_i$, temporal anchor $\tau_i$, source provenance $\pi_i$, associated entities $U_i$, and applicability scope $\Omega_i$. The aggregated reliability weight $\rho_i^t\in[0,1]$ summarizes the relative contribution of accumulated evidence sources, while $s_i^t$ denotes the memory's lifecycle state.

Adaptation therefore serves as the interface between heterogeneous memory storage and lifecycle governance: it preserves evidence traceability while exposing the structural information required for later validity estimation and conflict\hyp{}aware reasoning. Newly created cells are initialized as \textsc{Candidate}, with their validity and lifecycle states updated through subsequent evidence\hyp{}based governance. This process preserves traceable evidence without prematurely resolving conflicts or making query\hyp{}specific admission decisions. Detailed adaptation rules, evidence alignment, and source reliability weights are provided in Appendix~\ref{app:memory-adaptation}. These operations establish the evidence structure required for downstream lifecycle governance while preserving the original backend information.

\subsection{Stage II: Lifecycle\hyp{}Aware Memory Governance}
\label{sec:calm-lifecycle-governance}

After Stage I, CALM preserves evidence provenance and links evolving claims through a lifecycle graph. Identical normalized claims are consolidated at the memory\hyp{}cell level and their evidence counts are updated; the current implementation does not assign an independent support\hyp{}edge score. Conflict and dependency links are identified by deterministic rules:
\begin{equation}
\begin{aligned}
\mathcal{G}_t
&=(\mathcal{M}_t,\mathcal{E}_{\mathrm{conf}}^t,
                 \mathcal{E}_{\mathrm{dep}}^t),\\
p_{ij}^{r,t}&=f_r(m_i^t,m_j^t),
\quad r\in\{\mathrm{conf},\mathrm{dep}\}.
\end{aligned}
\label{eq:calm-lifecycle-graph}
\end{equation}

Conflict links are retained for detected opposing claims. A non\hyp{}weak conflict with strength at least $0.75$ can trigger supersession; weaker conflicts are recorded without automatically replacing either claim. A dependency edge $(j,i)$ points from the source or prerequisite memory $m_j$ to the dependent memory $m_i$. Dependency links are created when the relevant rule matches, without an additional global edge threshold. Full rule assignments are given in Appendix~\ref{app:lifecycle-governance}.

CALM tracks accumulated support $S_i^t$, conflict risk $C_i^t$, and invalid\hyp{}dependency risk $D_i^t$ using retained evidence and lifecycle events. In the implementation described in Appendix~\ref{app:lifecycle-governance}, these quantities support governance decisions; they are not fed into a separately recomputed sigmoid validity score. Instead, memory validity is maintained as a stored operational attribute alongside the persistent lifecycle state:
\begin{equation}
\begin{aligned}
(V_i^t,s_i^t)
&=\mathcal{U}_{\mathrm{life}}\!\Big(
V_i^{t-1},s_i^{t-1},\Delta\mathcal{E}_i^t,\\
&\qquad\Delta\mathcal{E}_{\mathrm{conf}}^t,
\Delta\mathcal{E}_{\mathrm{dep}}^t
\Big).
\end{aligned}
\label{eq:calm-lifecycle-update}
\end{equation}

Here $\mathcal U_{\mathrm{life}}$ denotes deterministic lifecycle and stored\hyp{}validity updates, not a learned scoring function. Direct user evidence or confidence of at least $0.60$ activates a candidate, while repeated support or confidence of at least $0.80$ can promote it to \textsc{Stable}. Strong supersession caps the previous memory's stored validity at $0.15$. Temporal information identifies the order and applicability intervals of updates; elapsed time alone does not trigger a \textsc{Stale} transition. Dependency invalidation can trigger verification or make a dependent memory \textsc{Stale}, while archival is governed by an optional retention policy. These persistent updates are distinct from Stage\hyp{}III query\hyp{}time eligibility decisions; details appear in Appendix~\ref{app:lifecycle-governance}.

\subsection{Stage III: Query\hyp{}Conditioned Memory Gating}
\label{sec:calm-stage3}

Query\hyp{}independent validity does not guarantee eligibility for a
specific query. A valid memory may fall outside the query's temporal
scope or applicability conditions, while a superseded memory may
remain useful for historical queries. CALM therefore governs
query\hyp{}time admission over retrieved memories rather than relying
on semantic relevance alone.

For a query $q$, let $\mathcal{R}_q\subseteq\mathcal{M}_t$ denote the
candidate set provided by the upstream retriever. For each retrieved
memory $m_i^t\in\mathcal{R}_q$, CALM constructs the four\hyp{}dimensional
representation
\begin{equation}
\mathbf{x}_{q,i}^{t}
=\left[\zeta_{q,i}^{t},V_i^{t},\psi_{q,i}^{t},1-R_{q,i}^{t}\right],
\label{eq:calm-main-gate-features}
\end{equation}
where $\zeta_{q,i}^{t}$ denotes semantic relevance, $V_i^{t}$ is the
query\hyp{}independent validity estimated in Stage II, $\psi_{q,i}^{t}$
measures query--memory compatibility, and $R_{q,i}^{t}$ summarizes
query\hyp{}conditioned conflict, dependency, temporal, and lifecycle risk.

CALM applies either a fixed training\hyp{}free policy (CALM\hyp{}ZS) or a
learnable policy (CALM) to the same features:
\begin{equation}
F_b(\mathbf{x}_{q,i}^{t})=
\begin{cases}
\mathbf{w}_{\mathrm{zs}}^{\top}\mathbf{x}_{q,i}^{t},
 & b=\mathrm{zs},\\
\sigma\!\left(f_{\phi}(\mathbf{x}_{q,i}^{t})\right),
 & b=\mathrm{learn},
\end{cases}
\label{eq:calm-main-gate-policy}
\end{equation}
where $\mathbf{w}_{\mathrm{zs}}=(1/4,1/4,1/4,1/4)^{\top}$ is fixed,
and $f_{\phi}$ is trained only on development instances with
independently constructed query--memory supervision.

Both policies first apply the same hard eligibility filter
$h_{q,i}^{t}\in\{0,1\}$, defined in Appendix~\ref{app:calm-stage3}. The final binary
admission decision is
\begin{equation}
g_{q,i}^{b,t}=h_{q,i}^{t}\,
\mathbb{I}\!\left[
F_b(\mathbf{x}_{q,i}^{t})\geq\theta_b
\right],
\label{eq:calm-main-gate-admission}
\end{equation}
where $\theta_{\mathrm{zs}}$ is the fixed CALM\hyp{}ZS threshold and
$\theta_{\mathrm{learn}}=0.50$ is the learned\hyp{}gate threshold.
Their values and implementation details are specified in
Appendix~\ref{app:calm-stage3}. The working context is
\begin{equation}
\mathcal{C}_q^b=
\{m_i^t\in\mathcal{R}_q\mid g_{q,i}^{b,t}=1\}.
\label{eq:calm-main-gate-context}
\end{equation}
Admitted memories are ordered by their policy scores and packed
under the same memory\hyp{}token budget. Rejected memories remain in
persistent storage and can support compatible historical queries.

\section{Experiments}

\subsection{Experimental Setup}

\begin{table*}[t]
\centering

{
% Local commands used only in this table
\newcommand{\best}[1]{\textbf{#1}}
\newcommand{\second}[1]{\underline{#1}}
\newcommand{\methodrow}{\rowcolor[gray]{0.92}}

\normalsize
\setlength{\tabcolsep}{1.5mm}

\begin{tabularx}{\textwidth}{
    l
    *{7}{>{\centering\arraybackslash}X}
}
\toprule
\textbf{Method}
& \textbf{Overall}
& \textbf{SS-User}
& \textbf{SS-Pref.}
& \textbf{SS-Asst.}
& \textbf{K-Upd.}
& \textbf{Multi}
& \textbf{Temp.}
\\
\midrule

% ==================== Qwen3 ====================
\multicolumn{8}{c}{
    \includegraphics[height=0.95em]{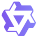}
    \hspace{0.25em}
    \textbf{Qwen3-30B-A3B-Instruct-2507}
}
\\
\midrule

Mem0 & 40.20 & 75.71 & 50.00 & 10.71 & 44.87 & 41.35 & 27.82 \\

MemOS & 68.80 & 88.57 & 73.33 & 64.29 & 65.38 & \second{64.66} & \second{65.41} \\

EverMemOS & 66.40 & 88.57 & 50.00 & 60.71 & 87.18 & 62.41 & 52.63 \\

LightMem & 67.00 & \second{94.29} & \second{76.67} & 28.57 & 73.08 & \second{64.66} & \second{65.41} \\

\methodrow
\textbf{CALM-ZS} & \second{76.60} & 91.43 & \best{93.33} & \second{91.07} & \best{93.59} & 60.15 & \second{65.41} \\

\methodrow
\textbf{CALM} & \best{79.60} & \best{95.71} & \best{93.33} & \best{92.86} & \second{91.03} & \best{66.92} & \best{68.42} \\

\midrule

% ==================== GPT-4.1-mini ====================
\multicolumn{8}{c}{
    \includegraphics[height=0.95em]{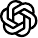}
    \hspace{0.25em}
    \textbf{GPT-4.1-mini}
}
\\
\midrule

Mem0 & 70.80 & 87.14 & 93.33 & 39.29 & 71.79 & 66.92 & 73.68 \\

MemOS & 78.80 & \second{94.29} & \best{96.67} & 69.64 & 78.21 & 71.43 & 78.20 \\

EverMemOS & \best{83.00} & \best{97.14} & \second{93.33} & \second{85.71} & \second{89.74} & \best{73.68} & 77.44 \\

LightMem & 69.60 & 90.00 & 76.67 & 23.21 & 88.46 & 51.13 & \best{84.21} \\

\methodrow
\textbf{CALM-ZS} & \second{79.40} & 88.57 & 86.67 & 83.93 & \second{89.74} & 67.67 & 76.69 \\

\methodrow
\textbf{CALM} & \best{83.00} & 94.29 & 83.33 & \best{89.29} & \best{94.87} & \second{72.18} & \second{78.20} \\

\midrule

% ==================== GPT-4o-mini ====================
\multicolumn{8}{c}{
    \includegraphics[height=0.95em]{Figures/gpt.pdf}
    \hspace{0.25em}
    \textbf{GPT-4o-mini}
}
\\
\midrule

Mem0 & 66.40 & 82.86 & 90.00 & 26.79 & 66.67 & 63.16 & 72.18 \\

MemOS & \second{77.80} & \best{95.71} & \best{96.67} & 67.86 & 74.36 & \second{70.68} & \best{77.44} \\

EverMemOS & 65.40 & \second{91.43} & \second{93.33} & 55.36 & 67.95 & 60.15 & 53.38 \\

LightMem & 69.60 & 87.14 & 66.67 & 32.14 & \best{83.33} & \best{71.43} & 66.92 \\

\methodrow
\textbf{CALM-ZS} & 75.40 & \second{91.43} & 70.00 & \second{71.43} & \second{82.05} & 69.17 & 72.18 \\

\methodrow
\textbf{CALM} & \best{78.20} & \best{95.71} & 80.00 & \best{80.36} & 80.77 & 68.42 & \second{75.94} \\

\bottomrule
\end{tabularx}
}

\caption{Main results on LongMemEval. Accuracy (\%) by method across
question types. SS denotes single-session, Pref. preference, Asst. assistant,
K-Upd. knowledge-update, Multi multi-session, and Temp. temporal reasoning.
The best and second-best results within each backbone are shown in bold and
underlined, respectively.}
\label{tab:longmemeval-main}
\end{table*}

\subsubsection{Datasets and Baselines}
We evaluate CALM on LongMemEval\hyp{}S~\citep{longmemeval}, which contains 500 questions across six memory\hyp{}reasoning categories, and LoCoMo~\citep{locomo}, which contains 1,986 questions over ten long conversations. Our reported LoCoMo evaluation covers the 1,540 non\hyp{}adversarial questions; the remaining 446 adversarial questions are excluded.
All methods are evaluated on the corresponding benchmark conversations and question subsets. External baselines retain their native memory\hyp{}construction and retrieval pipelines. For CALM, benchmark\hyp{}provided LoCoMo observations and summaries are incorporated as alternative representations of the same conversation evidence, not as answer labels or gate supervision. We compare with Mem0~\citep{mem0}, MemOS~\citep{memos}, EverMemOS~\citep{evermemos}, and LightMem~\citep{lightmem}. Detailed protocols are given in the Appendix~\ref{sec:datasets_baseline}.

\begin{table}[t]
\centering

% Local commands used only in this table
\newcommand{\best}[1]{\textbf{#1}}
\newcommand{\second}[1]{\underline{#1}}
\newcommand{\methodrow}{\rowcolor[gray]{0.92}}

{
\small
\setlength{\tabcolsep}{0.7mm}

\begin{tabularx}{\columnwidth}{
    l
    *{5}{>{\centering\arraybackslash}X}
}
\toprule
\textbf{Method}
& \textbf{Overall}
& \mbox{\textbf{S-Hop}}
& \mbox{\textbf{M-Hop}}
& \textbf{Open}
& \textbf{Temp.}
\\
\midrule

% ==================== Qwen3 ====================
\multicolumn{6}{l}{\textit{\textbf{Qwen3-30B-A3B-Instruct-2507}}} \\

Mem0 & 59.09 & 65.16 & 54.26 & 45.83 & 51.40 \\
MemOS & 76.04 & 81.81 & 73.76 & 59.38 & 67.91 \\
EverMemOS & \second{87.99} & \second{92.75} & \best{86.17} & \second{70.83} & \second{82.24} \\
LightMem & 72.01 & 78.60 & 72.34 & 57.29 & 58.88 \\

\methodrow
\textbf{CALM-ZS} & 85.78 & 92.15 & \second{84.04} & 67.71 & 76.01 \\

\methodrow
\textbf{CALM} & \best{88.70} & \best{94.65} & 82.27 & \best{73.96} & \best{83.18} \\

\midrule

% ==================== GPT-4.1-mini ====================
\multicolumn{6}{l}{\textit{\textbf{GPT-4.1-mini}}} \\

Mem0 & 64.22 & 68.97 & 61.70 & 50.00 & 58.26 \\
MemOS & 80.84 & 85.37 & 79.43 & 64.58 & 75.08 \\
EverMemOS & \second{93.05} & \second{96.67} & \second{91.84} & \best{76.04} & \second{89.72} \\
LightMem & 78.90 & 82.40 & 73.40 & 51.04 & 82.87 \\

\methodrow
\textbf{CALM-ZS} & 90.13 & 95.36 & 86.17 & 71.88 & 85.36 \\

\methodrow
\textbf{CALM} & \best{93.51} & \best{97.03} & \best{93.26} & \second{73.96} & \best{90.34} \\

\midrule

% ==================== GPT-4o-mini ====================
\multicolumn{6}{l}{\textit{\textbf{GPT-4o-mini}}} \\

Mem0 & 61.17 & 66.71 & 58.16 & 40.62 & 55.45 \\
MemOS & 75.97 & 81.45 & 69.15 & 60.42 & 72.27 \\
EverMemOS & 86.75 & \second{91.08} & 86.17 & \best{66.67} & 81.93 \\
LightMem & 72.99 & 76.81 & 67.02 & 45.83 & 76.32 \\

\methodrow
\textbf{CALM-ZS} & \second{86.82} & 89.18 & \second{86.88} & 62.50 & \best{87.85} \\

\methodrow
\textbf{CALM} & \best{88.70} & \best{91.79} & \best{89.01} & \second{64.58} & \second{87.54} \\

\bottomrule
\end{tabularx}
}

\caption{Main results on LoCoMo.}
\label{tab:locomo_results}
\end{table}

\subsubsection{Metrics and implementation details} We evaluate our method from three dimensions: answer effectiveness, memory selection quality, and query\hyp{}time efficiency. For answer effectiveness, we report accuracy on LongMemEval\hyp{}S and LoCoMo using Qwen3-30B\hyp{}A3B\hyp{}Instruct-2507~\citep{qwen3}, GPT-4.1-mini, and GPT-4o\hyp{}mini as answer models. Outputs from Qwen3 are evaluated by qwen\hyp{}turbo, while those from GPT-4.1-mini and GPT-4o\hyp{}mini~\citep{gpt4} are evaluated by their corresponding models. Memory selection quality is assessed using erroneous\hyp{}memory blocking (EMB), useful\hyp{}memory retention (UMR), and conflict leakage (CL), together with erroneous\hyp{}memory injection and historical\hyp{}reconstruction analyses. Efficiency is evaluated through accuracy under different memory\hyp{}token budgets and a query\hyp{}time breakdown of context\hyp{}token usage and latency. Detailed metric definitions, evidence\hyp{}coverage diagnostics, prompts, cost accounting, and training configurations are provided in the Appendix~\ref{sec:evaluation-protocol}.

\subsection{Main Results}
\subsubsection{Main results on LongMemEval\hyp{}S}
% \begin{table}[t]
% \centering

% % Local commands used only in this table
% \newcommand{\best}[1]{\textbf{#1}}
% \newcommand{\second}[1]{\underline{#1}}

% {
% \small
% \setlength{\tabcolsep}{6pt}

% \begin{tabular}{@{}lccc@{}}
% \toprule
% \textbf{Variant}
% & \textbf{EMB $\uparrow$}
% & \textbf{UMR $\uparrow$}
% & \textbf{CL $\downarrow$}
% \\
% \midrule

% No-gating & 0.00 & \best{100.00} & 100.00 \\

% \textbf{CALM-ZS} & \second{79.32} & 67.76 & \second{21.68} \\

% \textbf{CALM} & \best{81.30} & \second{72.06} & \best{20.00} \\

% \bottomrule
% \end{tabular}
% }

% \caption{Conflict-blocking performance.}
% \label{tab:conflict_blocking}
% \end{table}

\begin{table}[t]
\centering

% Local commands used only in this table
\newcommand{\best}[1]{\textbf{#1}}
\newcommand{\second}[1]{\underline{#1}}

{
\small
\setlength{\tabcolsep}{6pt}

\begin{tabular*}{\columnwidth}{
@{}l@{\extracolsep{\fill}}ccc@{}
}
\toprule
\textbf{Variant}
& \textbf{EMB $\uparrow$}
& \textbf{UMR $\uparrow$}
& \textbf{CL $\downarrow$}
\\
\midrule

No-gating & 0.00 & \best{100.00} & 100.00 \\

\textbf{CALM-ZS} & \second{79.32} & 67.76 & \second{21.68} \\

\textbf{CALM} & \best{81.30} & \second{72.06} & \best{20.00} \\

\bottomrule
\end{tabular*}
}

\caption{Conflict-blocking performance.}
\label{tab:conflict_blocking}
\end{table}

Table~\ref{tab:longmemeval-main} shows that CALM achieves the best or tied\hyp{}best overall accuracy across all three backbones, reaching 79.60\%, 83.00\%, and 78.20\% with Qwen3-30B, GPT-4.1-mini, and GPT-4o\hyp{}mini, respectively. CALM also maintains competitive performance across different question types and consistently achieves the best results on SS\hyp{}Asst., indicating stable generalization across backbones and memory scenarios. Moreover, its consistent gains over CALM\hyp{}ZS demonstrate that learnable query\hyp{}conditioned gating improves memory selection by preventing irrelevant or conflicting records from entering the reasoning context.

\subsubsection{Main Results on LoCoMo.}
\label{sec:main-locomo}

% As shown in Table~\ref{tab:locomo_results}, CALM achieves the best
% overall accuracy across all three answer models, reaching
% 88.70\%, 93.51\%, and 88.70\%, respectively. Compared with CALM-ZS,
% the learned admission module improves performance by 2.92, 3.38, and
% 1.88 percentage points, indicating that query-conditioned admission
% provides benefits beyond fixed filtering strategies.

% CALM remains competitive across single-hop, multi-hop, open-domain, and
% temporal questions. Together with the ablations and simple-alternative
% comparisons, these results indicate that the gains come from the
% combination of lifecycle-aware governance and learned admission rather
% than simple post-retrieval heuristics. The lower Qwen3-30B multi-hop
% score suggests a potential trade-off between stronger conflict
% suppression and preserving evidence for complex reasoning.

As shown in Table~\ref{tab:locomo_results}, CALM achieves the best overall accuracy across all three answer models, reaching 88.70\%, 93.51\%, and 88.70\%, respectively. Compared with CALM\hyp{}ZS, the learned admission module improves performance by 2.92, 3.38, and 1.88 percentage points, showing that query\hyp{}conditioned admission provides benefits beyond fixed filtering. CALM remains competitive across single\hyp{}hop, multi\hyp{}hop, open\hyp{}domain, and temporal questions. Together with ablations and simple\hyp{}alternative comparisons, these results indicate that the gains arise from combining lifecycle\hyp{}aware governance with learned admission rather than simple post\hyp{}retrieval heuristics. The lower Qwen3-30B multi\hyp{}hop score suggests a trade\hyp{}off between stronger conflict suppression and preserving evidence for complex reasoning.

\subsubsection{Conflict Blocking Analysis.} We evaluate the gating module using EMB, UMR, and CL, as reported in Table~\ref{tab:conflict_blocking}. Without gating, all retrieved memories are admitted into the reasoning context, trivially yielding 100.00\% UMR but resulting in 0.00\% EMB and 100.00\% CL. Both CALM variants substantially improve this trade\hyp{}off. CALM\hyp{}ZS blocks 79.32\% of erroneous memories and reduces CL to 21.68\%, while retaining 67.76\% of useful memories. The learned CALM gate further increases EMB and UMR to 81.30\% and 72.06\%, respectively, while reducing CL to 20.00\%. These results show that learned gating more effectively suppresses conflicting memories while preserving useful evidence.

\subsection{Ablation Study}
\begin{table}[t]
\centering

{
\small
\setlength{\tabcolsep}{5pt}
\renewcommand{\arraystretch}{0.95}

\begin{tabular*}{\columnwidth}{
@{}l@{\extracolsep{\fill}}c@{}
}
\toprule
\textbf{Variant}
& \textbf{Overall Acc. ($\uparrow$)} \\
\midrule

\multicolumn{2}{@{}l}{\textit{Stage-wise ablations}} \\
w/o Stage I & 69.80 \\
w/o Stage II & 75.60 \\
w/o Stage III & 74.00 \\

\midrule

\multicolumn{2}{@{}l}{\textit{Simple filtering and reranking}} \\

Temporal-Filter & 74.40 \\
Recency-Rerank & 75.60 \\
Validity-Rerank & 77.20 \\

\midrule

CALM-ZS & 76.60 \\
\textbf{CALM} & \textbf{79.60} \\

\bottomrule
\end{tabular*}
}

\caption{Stage-wise ablations and simple alternatives on
LongMemEval-S with Qwen3-30B-A3B-Instruct-2507.
}
\label{tab:stage_simple}
\end{table}
% All variants use the same retrieved candidates and memory-token budget.

\subsubsection{Ablation Analysis of CALM}

% Table~\ref{tab:stage_simple} examines the contributions of the three stages. Removing Stages I, II, and III reduces overall accuracy from 79.60\% to 69.80\%, 75.60\%, and 74.00\%, respectively, supporting their complementary roles. The component ablations in Table~5 further show that excluding the stored-validity feature reduces knowledge-update accuracy from 91.03\% to 87.18\%, while removing temporal cues lowers temporal-reasoning accuracy from 68.42\% to 60.15\%.

% The three-stage ablation in Table~\ref{tab:stage_simple} examines the contributions of each component. Removing Stages I, II, and III reduces overall accuracy from 79.60\% to 69.80\%, 75.60\%, and 74.00\%, respectively, supporting their complementary roles. Component ablations further show that excluding the stored-validity feature reduces knowledge-update accuracy from 91.03\% to 87.18\%, while removing temporal cues lowers temporal-reasoning accuracy from 68.42\% to 60.15\%.

% The three-stage ablation in Table~\ref{tab:stage_simple} examines the role of each component. Removing Stages I, II, and III substantially reduces overall accuracy, supporting their complementary roles. Component ablations further show that stored validity is important for knowledge updates, while temporal cues are crucial for temporal reasoning.

The three\hyp{}stage ablation in Table~\ref{tab:stage_simple} examines the role of each component. Removing Stages I, II, and III substantially reduces overall performance, supporting their complementary roles in memory governance. Component ablations further show that stored validity is important for knowledge updates, while temporal cues are crucial for temporal reasoning.

\paragraph{Comparison with Simple Alternatives.}
To examine whether CALM's gains stem from simple post\hyp{}retrieval heuristics, we compare temporal filtering, recency reranking, and validity reranking under identical memory representations, retrieved candidates, answer model, evaluation protocol, and memory\hyp{}token budget (Appendix~\ref{sec:detailed-ablation}). Temporal filtering excludes time\hyp{}incompatible records, while reranking adjusts candidate ordering without explicit admission. As shown in Table~\ref{tab:stage_simple}, Validity\hyp{}Rerank 77.20\% outperforms the temporal alternatives and CALM\hyp{}ZS
76.60\%, but trails CALM 79.60\% by 2.40\%. This gain supports CALM's complete filtering\hyp{}and\hyp{}admission policy rather than the learned predictor alone. Moreover, CALM and CALM\hyp{}ZS share identical hard filters, features, and context\hyp{}packing rules, isolating a 3.00\% improvement from learned eligibility scoring.

\begin{figure}[t]
    \centering
    \includegraphics[width=\linewidth]{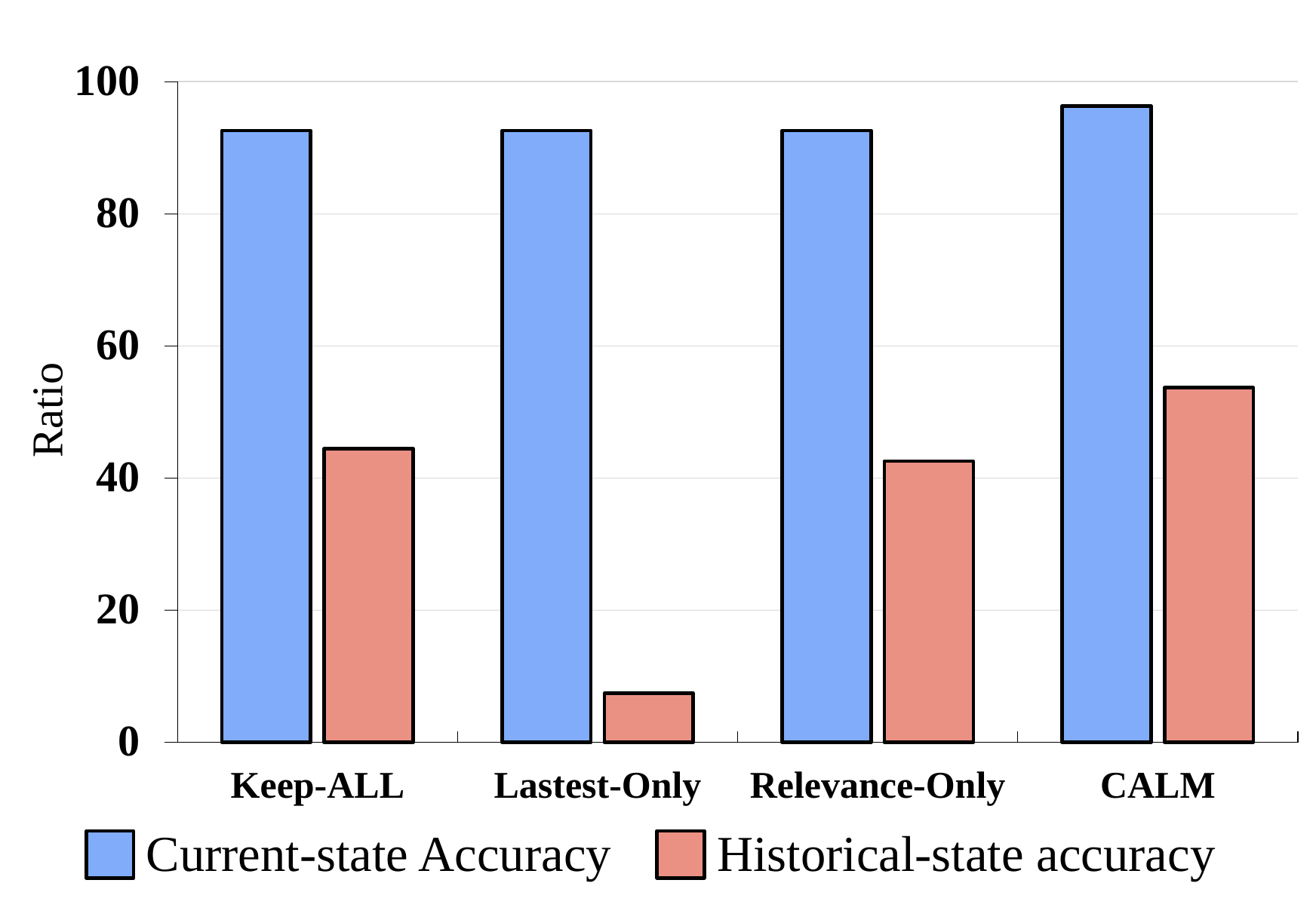}
    \caption{Performance of memory management strategies in reconstructing current and historical states.}
    \label{fig:historical_reconstruction}
\end{figure}

\subsubsection{Memory Retention and Conflict Robustness}
Detailed ablations and erroneous\hyp{}memory injection results are provided in the appendix~\ref{sec:memory-conflict-robustness}. Additional sensitivity analysis of source reliability weighting is provided in Appendix~\ref{app:weight-sensitivity}. The analysis shows that lifecycle governance first reduces uncertainty in persistent memory by identifying outdated or conflicting states, while query\hyp{}conditioned gating further prevents such memories from contaminating the working context. This separation allows CALM to remain robust as memory noise increases, preserving useful evidence while filtering most misleading records.
\begin{table}[t]
\centering

% Local commands used only in this table
\newcommand{\best}[1]{\textbf{#1}}
\newcommand{\second}[1]{\underline{#1}}

{
\small
\setlength{\tabcolsep}{6pt}

\begin{tabular}{@{}lccc@{}}
\toprule
\textbf{Variant}
& \textbf{Overall $\uparrow$}
& \textbf{K-Upd. $\uparrow$}
& \textbf{Temp. $\uparrow$}
\\
\midrule

\textit{w/o Conflict}
& 73.00
& \second{89.74}
& 62.41
\\

\textit{w/o Temporal}
& 72.60
& 88.46
& 60.15
\\

\textit{w/o Validity}
& 72.40
& 87.18
& 63.16
\\

\textit{w/o Compatibility}
& 74.20
& \second{89.74}
& 63.91
\\

\textit{w/o Risk}
& \second{75.20}
& \second{89.74}
& \second{64.66}
\\

\textbf{CALM}
& \best{79.60}
& \best{91.03}
& \best{68.42}
\\

\bottomrule
\end{tabular}
}

\caption{Lifecycle and gating ablation results.}
\label{tab:lifecycle_gating_ablation}
\end{table}
\subsubsection{Historical Reconstruction}
The historical\hyp{}reconstruction ablation evaluates whether superseded memories can be retained without interfering with current\hyp{}state reasoning. As shown in Figure~\ref{fig:historical_reconstruction}, Latest\hyp{}Only performs poorly because earlier states are discarded, while Keep\hyp{}All and Relevance\hyp{}Only recover only part of the historical evidence. CALM achieves the best accuracy on both current- and historical\hyp{}state queries, surpassing the strongest baseline by 3.71\% and 9.26\%, respectively. These results confirm that CALM effectively separates memory retention from query\hyp{}time use.
\begin{figure}[t]
    \centering
    \includegraphics[width=\linewidth]{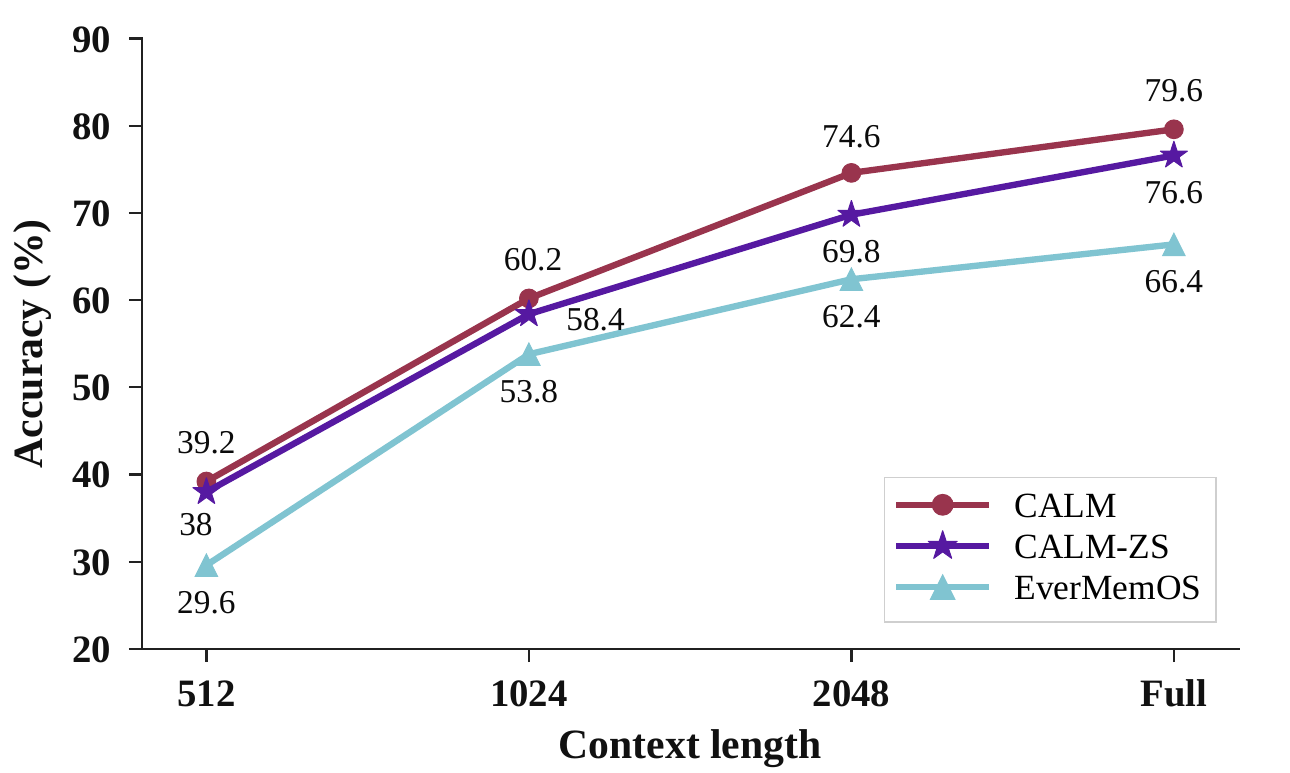}
    \caption{Accuracy across memory\hyp{}token budgets.}
    \label{fig:context_budget_scaling}
\end{figure}

\subsection{Efficiency Analysis}
\subsubsection{Context Budget Scaling}
As shown in Figure~\ref{fig:context_budget_scaling}, accuracy improves with the memory budget across all methods, while CALM consistently performs best across evaluated budget settings. Its largest gain over CALM\hyp{}ZS occurs at 2048 tokens, reaching 4.8\%, indicating that learned gating is especially effective under tighter context constraints. Although the gap narrows as more evidence is retained, CALM remains superior under the full\hyp{}context setting, demonstrating a better accuracy--context efficiency trade\hyp{}off with robust context utilization.

\subsubsection{Inference\hyp{}Time Cost Analysis}

The learned gating mechanism improves accuracy with only moderate inference overhead. CALM gains 3.0\% over CALM\hyp{}ZS while increasing memory usage by 3.1\%, as shown in Table~\ref{tab:query_time_efficiency}. The additional cost is concentrated in gating: retrieval latency remains unchanged, answer\hyp{}generation latency increases by only 0.05 seconds, and total latency rises from 12.00 to 12.84 seconds. CALM also substantially outperforms EverMemOS in both accuracy and latency. These results indicate that the performance gain primarily comes from more effective memory selection rather than larger contexts or more expensive generation. Here, R/G denotes retrieval and gating latency.

% \begin{table}[t]
% \centering

% % Local commands used only in this table
% \newcommand{\best}[1]{\textbf{#1}}
% \newcommand{\second}[1]{\underline{#1}}

% {
% \small
% \setlength{\tabcolsep}{1mm}

% \begin{tabularx}{\columnwidth}{
%     l
%     *{5}{>{\centering\arraybackslash}X}
% }
% \toprule
% \textbf{Method}
% & \textbf{Acc. $\uparrow$}
% & \textbf{Mem. Tok. $\downarrow$}
% & \textbf{R/G (s) $\downarrow$}
% & \textbf{Ans. (s) $\downarrow$}
% & \textbf{Total (s) $\downarrow$}
% \\
% \midrule

% EverMemOS
% & 66.20
% & \best{2775.43}
% & 46.05 / --
% & 14.39
% & 61.65
% \\

% \textbf{CALM-ZS}
% & \second{76.60}
% & \second{3850.00}
% & \best{6.02} / \best{0.40}
% & \best{1.44}
% & \best{12.00}
% \\

% \textbf{CALM}
% & \best{79.60}
% & 3970.81
% & \best{6.02} / \second{1.03}
% & \second{1.49}
% & \second{12.84}
% \\

% \bottomrule
% \end{tabularx}
% }

% \caption{Results on query-time efficiency.}
% \label{tab:query_time_efficiency}
% \end{table}

\begin{table}[t]
\centering

{
\small
\setlength{\tabcolsep}{2pt}

% Local commands
\newcommand{\best}[1]{\textbf{#1}}
\newcommand{\second}[1]{\underline{#1}}

\begin{tabular*}{\columnwidth}{
@{}l@{\extracolsep{\fill}}ccccc@{}
}
\toprule
\textbf{Method}
& \mbox{\textbf{Acc.}$\uparrow$}
& \mbox{\textbf{Tok.}$\downarrow$}
& \mbox{\textbf{R/G}$\downarrow$}
& \mbox{\textbf{Ans.}$\downarrow$}
& \mbox{\textbf{Total}$\downarrow$}
\\
\midrule

EverMemOS
& 66.40
& \best{2775}
& \mbox{46.05/--}
& 14.39
& 61.65
\\

\textbf{CALM-ZS}
& \second{76.60}
& \second{3850}
& \mbox{\best{6.02}/\best{0.40}}
& \best{1.44}
& \best{12.00}
\\

\textbf{CALM}
& \best{79.60}
& 3971
& \mbox{\best{6.02}/\second{1.03}}
& \second{1.49}
& \second{12.84}
\\

\bottomrule
\end{tabular*}
}

\caption{Results on query-time efficiency.}
\label{tab:query_time_efficiency}
\end{table}

\section{Conclusion}
Reliable long\hyp{}horizon memory requires more than retaining information: it requires deciding when preserved evidence should inform an answer. We presented CALM, a framework that separates evidence\hyp{}grounded, query\hyp{}independent lifecycle governance from query\hyp{}conditioned admission of retrieved memories. By preserving provenance and superseded states while controlling admission through query compatibility and risk, CALM supports historical reconstruction without compromising current\hyp{}state reasoning. Experiments on LongMemEval\hyp{}S and LoCoMo across Qwen and GPT backbones show the best overall accuracy. Conflict\hyp{}blocking, historical\hyp{}reconstruction, and comparisons with simple filtering and reranking strategies further demonstrate the value of governing memory use beyond semantic relevance alone. Future work will explore scalable lifecycle maintenance in dynamic, multimodal environments.

\section*{Limitations}
CALM has several limitations. First, its lifecycle governance relies on deterministic rules and available evidence metadata, which may miss implicit or complex conflicts. Second, query\hyp{}conditioned gating depends on upstream retrieval quality and may occasionally exclude useful evidence, particularly for multi\hyp{}hop reasoning. Third, the learned gate requires development\hyp{}set supervision, and its generalization to unseen domains remains to be explored. Finally, our evaluation is limited to LongMemEval\hyp{}S and the non\hyp{}adversarial subset of LoCoMo. The scalability of lifecycle\hyp{}graph maintenance in larger, continuously evolving memory environments also warrants further investigation.


\bibliography{custom}

\appendix
\section{Additional Method Details}
\label{sec:supp-methods}

This section complements Section~3 of the main paper and reports the implementation details omitted from the main text. We retain the same three stages and describe the operational realization of evidence adaptation, relation estimation, lifecycle transitions, query\hyp{}conditioned compatibility and risk computation, gate supervision, and online context construction. Unless otherwise stated, the interaction\hyp{}step superscript $t$ is omitted for readability.

\subsection{Stage I: Evidence\hyp{}Grounded Memory Adaptation}
\label{app:memory-adaptation}

\begin{table}[t]
\centering
\small
\caption{Source reliability weights used for evidence aggregation.
The weights are fixed heuristic values determined by evidence provenance
and granularity. They are not calibrated probabilities of factual
correctness and are not tuned on evaluation instances.}
\label{tab:source-weight}
\begin{tabular}{lll}
\toprule
Dataset & Evidence source & Weight $\nu$ \\
\midrule
\multirow{3}{*}{LongMemEval-S}
& Raw dialogue span & 0.74 \\
& Rule-derived span & 0.82 \\
& Session anchor & 0.62 \\
\midrule
\multirow{4}{*}{LoCoMo}
& Dialogue turn & 0.72 \\
& Observation & 0.95 \\
& Event summary & 0.88 \\
& Session summary & 0.82 \\
\bottomrule
\end{tabular}
\end{table}

\paragraph{Input records and adaptation granularity.}
The adapter processes records exposed by the existing memory backend and preserves their original content and provenance. Structured attributes are mapped to corresponding cell fields when available. For textual records, deterministic rules extract explicit facts or bounded spans; when reliable decomposition is unavailable, the original source span or dialogue turn is retained as a single cell. Therefore, a memory cell is the unit of evidence and lifecycle management, but is not necessarily an atomic proposition. Our implementation uses deterministic adaptation rules for reproducibility, while the representation remains backend\hyp{}agnostic and independent of any specific extraction model.

On LongMemEval\hyp{}S, raw messages are segmented at punctuation\hyp{}aware boundaries into spans of at most 900 characters, with at most 12 spans retained per message. Rule\hyp{}derived spans covering explicit updates, temporal statements, preferences, and cross\hyp{}session cues are capped at 220 characters. A session anchor additionally retains the first three messages of each session. For each evaluation question, memory construction only uses sessions available before or at the question date, preventing future information leakage.

On LoCoMo, each timestamped dialogue turn is preserved as a source\hyp{}level cell. Dataset\hyp{}provided observations, event summaries, and session summaries are incorporated as additional evidence representations. These records are available to all CALM variants and are not used as answer supervision, gate labels, or reference information. They are provided by the benchmark rather than generated by the evaluated model. Since multiple representations may describe the same conversation, they are treated as alternative evidence views rather than independent corroboration.

\paragraph{Cell attributes and missing information.}
The attribute $\kappa_i$ records a backend\hyp{}provided semantic category or a rule\hyp{}matched category (e.g., preference, profile attribute, event, constraint, or temporal fact). Entities $U_i$ and applicability scope $\Omega_i$ are populated only from available structured fields or explicit textual cues. Source provenance $\pi_i$ stores the record type and original interaction identifiers. The temporal anchor $\tau_i$ represents the available observation timestamp, while explicitly stated event times are stored separately. Missing attributes are left unspecified rather than inferred. These fields describe evidence properties but do not directly determine memory validity.

\paragraph{Evidence schema and provenance.}
Each memory cell follows Eq.~\eqref{eq:memory-cell}. Its evidence set contains items:
\begin{equation}
\label{eq:adaptation-evidence}
e=(\mathrm{id},u,\tau,\pi,r,\nu),
\end{equation}
where $\mathrm{id}$ identifies the evidence item, $u$ is the original source content, $\tau$ is the observation timestamp, $\pi$ stores provenance information including source type, session, turn, and speaker identity, $r$ distinguishes directly observed records from derived representations, and $\nu\in[0,1]$ denotes the assigned source reliability weight. Speaker identity is kept separate from evidence origin; for example, a LoCoMo participant is not treated as an assistant\hyp{}generated inference solely because the speaker label differs from \texttt{user}. Explicit event times are also distinguished from observation timestamps.

\paragraph{Evidence alignment and reliability.}
To identify exact duplicates, we normalize source text by lowercasing, trimming boundaries, and collapsing repeated whitespace. Records with identical normalized keys share a memory cell, and their evidence sets are merged by unique evidence identifiers while preserving original sources. Semantically similar records are not merged solely based on similarity. Records with the same entity and attribute but different values remain separate so that lifecycle governance can determine whether they represent conflicts or different temporal states.

We assign fixed source reliability weights $\nu$ according to evidence provenance and granularity, as summarized in Table~\ref{tab:source-weight}. These weights represent relative contributions during evidence aggregation rather than calibrated probabilities of factual correctness. Localized observations and explicit spans receive higher weights than broad summaries because they provide more direct evidence. The weights are fixed across datasets and models and are not tuned on evaluation instances.

To evaluate the sensitivity of this design choice, we compare alternative weighting strategies in Appendix~\ref{app:weight-sensitivity}. When multiple evidence items are aligned, the aggregated source reliability of a memory cell is computed as:
\begin{equation}
\rho_i^t=\max_{e\in\mathcal{E}_i^t}\nu(e).
\end{equation}

Stage II independently aggregates accumulated evidence and relation signals to estimate query\hyp{}independent validity $V_i^t$. Thus, $\rho_i^t$ only represents source\hyp{}level evidence weighting and should not be interpreted as a validity probability. Newly created cells are initialized as \textsc{Candidate}; subsequent promotion and lifecycle transitions are determined by Stage II rather than by the adaptation process itself.

\subsection{Source Reliability Weight Sensitivity}
\label{app:weight-sensitivity}

The source reliability weights in Table~\ref{tab:source-weight}
provide a lightweight evidence aggregation mechanism based on source
provenance and granularity. To examine whether CALM depends on this
specific weighting scheme, we compare three variants while keeping all
other components unchanged:

\begin{itemize}
    \item \textbf{Source\hyp{}specific}: the default setting using the weights
    in Table~\ref{tab:source-weight}.
    \item \textbf{Uniform}: all evidence sources are assigned the same
    weight $\nu=0.75$.
    \item \textbf{No weighting}: source distinctions are removed and
    all evidence items use $\nu=1$.
\end{itemize}

\begin{table}[t]
\centering
\small
\begin{tabular}{lc}
\toprule
Weighting Strategy & Accuracy (\%) \\
\midrule
No weighting ($\nu=1$) & 78.4 \\
Uniform weighting ($\nu=0.75$) & 78.9 \\
Source-specific weighting & 79.6 \\
\bottomrule
\end{tabular}
\caption{Sensitivity analysis of source reliability weighting on LongMemEval-S with Qwen3-30B-A3B-Instruct-2507.}
\label{tab:weight-sensitivity}
\end{table}

We evaluate these variants with Qwen3-30B\hyp{}A3B\hyp{}Instruct-2507
on LongMemEval\hyp{}S. The retrieved candidates, lifecycle
governance rules, gating policy, and answer\hyp{}generation configuration are
identical across variants. Results are shown in
Table~\ref{tab:weight-sensitivity}.

The results show that CALM remains effective under alternative weighting
strategies, while the default source\hyp{}specific weighting provides the best
performance. This indicates that CALM's improvements do not rely on a
particular manually assigned weight configuration; instead, the main
benefits arise from evidence\hyp{}grounded lifecycle governance and
query\hyp{}conditioned memory admission.

\subsection{Stage II: Lifecycle\hyp{}Aware Memory Governance}
\label{app:lifecycle-governance}

\paragraph{Relation identification.}
After Stage I, CALM constructs a lifecycle graph over canonical memory cells to represent supporting evidence, conflicting claims, and inter\hyp{}memory dependencies. Following the relation formulation in Eq.~(2) of the main paper, relation identification is implemented using deterministic rules based on evidence provenance, structured attributes, temporal information, and lexical overlap. No external LLM judge or learned pairwise relation classifier is used. The resulting confidence values represent operational relation strengths rather than calibrated probabilities.

The three relation types are instantiated differently. Support is handled through exact duplicate consolidation rather than a separately scored graph edge. Two memory records are merged when their contents match after lowercasing, trimming, and whitespace normalization. Their evidence records, source\hyp{}turn identifiers, and confidence information are retained or updated. Consequently, the implementation does not require a separate support\hyp{}edge threshold $\theta_{\mathrm{sup}}$.

For conflict and dependency relations, detected pairs receive rule\hyp{}specific strengths:
\begin{equation}
p_{ij}^{r,t}=f_r(m_i^t,m_j^t),
\qquad
r\in\{\mathrm{conf},\mathrm{dep}\}.
\label{eq:stageii_relation}
\end{equation}
The operational rules and decision criteria are summarized in Table~\ref{tab:relation_rules}.

Conflict identification first restricts candidate pairs using compatible entity--attribute slots or relevant personal claims. The detector then applies ordered rules for preference reversals, constraint changes, temporal location updates, task\hyp{}status changes, structured slot updates, and weaker opposing signals. Their confidence values are fixed by relation type, as reported in Table~\ref{tab:relation_rules}. All detected conflicts are retained, but only non\hyp{}weak conflicts with $p_{ij}^{\mathrm{conf}}\geq0.75$ trigger supersession. Thus, $0.75$ is a strong\hyp{}conflict decision threshold, not a minimum threshold for recording conflict relations. Conflict links are stored in both directions to preserve the relation for subsequent retrieval and consistency checks.

Dependency links follow the direction of evidence propagation. Specifically, an edge $(j,i)$ indicates that memory $m_j$ is a source or associated premise of memory $m_i$. Explicit source\hyp{}memory references receive the highest confidence, whereas weaker associations are determined by shared structured attributes, dialogue\hyp{}turn provenance, predefined semantic cues, or normalized token overlap. The overlap score is defined as
\begin{equation}
o_{ij}=
\frac{|T_i\cap T_j|}
{\sqrt{|T_i||T_j|}},
\end{equation}
where $T_i$ and $T_j$ are the sets of content tokens associated with the two memories. Dependency links are created when their corresponding rule conditions are satisfied; no additional global threshold $\theta_{\mathrm{dep}}$ is applied. The dependency strengths are subsequently used for invalidation\hyp{}risk propagation.

Importantly, semantic similarity alone does not establish equivalence between distinct claims. Non\hyp{}duplicate memories remain separate records, allowing the lifecycle graph to preserve conflicting statements, temporal updates, and independently supported information.

% % Requires booktabs and tabularx
% \providecommand{\CALMFill}[1]{%
%     \textbf{[VERIFY: #1]}%
% }

% \begin{table*}[t]
% \centering
% \small
% \setlength{\tabcolsep}{5pt}
% \renewcommand{\arraystretch}{1.1}

% \begin{tabularx}{\textwidth}{@{}lXXl@{}}
% \toprule

% \textbf{Relation}
% & \textbf{Evidence considered}
% & \textbf{Operational scoring rule}
% & \textbf{Threshold} \\

% \midrule

% Support
% & Aligned source claims and evidence provenance
% & \CALMFill{support scoring}
% & \CALMFill{theta sup} \\

% Conflict
% & Potentially incompatible claims, time scopes,
% and source evidence
% & \CALMFill{conflict scoring}
% & \CALMFill{theta conf} \\

% Dependency
% & Explicit premises and dependency cues
% between memory records
% & \CALMFill{dependency scoring}
% & \CALMFill{theta dep} \\

% \bottomrule
% \end{tabularx}

% \caption{
% Operational relation scoring rules and
% edge thresholds for lifecycle graph construction.
% }
% \label{tab:relation_rules}
% \end{table*}

% Requires booktabs and tabularx
\begin{table*}[t]
\centering
\small
\setlength{\tabcolsep}{4pt}
\renewcommand{\arraystretch}{1.18}

\begin{tabularx}{\textwidth}{@{}p{0.10\textwidth}>{\raggedright\arraybackslash}X>{\raggedright\arraybackslash}X>{\raggedright\arraybackslash}p{0.19\textwidth}@{}}
\toprule
\textbf{Relation}
& \textbf{Evidence considered}
& \textbf{Operational scoring rule}
& \textbf{Decision criterion} \\
\midrule

Support
& Exact matches after lowercasing and whitespace normalization; accumulated evidence and source confidence.
& Identical normalized records are merged, with evidence counts and confidence updated. No independent support-edge score $f_{\mathrm{sup}}$ is computed.
& Exact normalized match; no $\theta_{\mathrm{sup}}$. \\

\addlinespace[2pt]

Conflict
& Matching entity--attribute slots or compatible personal claims; preference polarity, constraints, location changes, and task status.
& $f_{\mathrm{conf}}=0.95$ (preference reversal), $0.94$ (temporal location change), $0.92$ (constraint change), $0.90$ (task-status change), $0.88$ (slot-value update), or $0.70$ (weak opposing signals with overlap $o\geq0.45$).
& All detected conflicts are linked. Strong conflict: $f_{\mathrm{conf}}\geq0.75$ and non-weak type. \\

\addlinespace[2pt]

Dependency
& Explicit source-memory IDs, shared structured slots, dialogue-turn provenance, predefined semantic cues, and token overlap.
& $f_{\mathrm{dep}}=0.95$ (explicit reference); $0.55$ (same slot, non-derived); $\max(0.90,o)$ (derived, same turn, $o\geq0.12$); $\max(0.80,o,s)$ (derived, semantic cue $s\geq0.75$); or $0.70$ (derived, topic overlap $o\geq0.45$).
& Link if a rule matches; no global $\theta_{\mathrm{dep}}$. Verify at cascade risk $D_i\geq0.70$; mark stale for superseded sources with $f_{\mathrm{dep}}\geq0.80$. \\

\bottomrule
\end{tabularx}

\par\vspace{3pt}
\begin{minipage}{\textwidth}
\footnotesize
\emph{Note.}
Token overlap is
$o=|T_i\cap T_j|/\sqrt{|T_i||T_j|}$,
where $T_i$ denotes the set of memory-content tokens.
The semantic association score $s$ is determined by predefined matching rules. Candidate memories become active with direct user evidence or confidence $\geq0.60$, and active memories become stable with at least two evidence items or confidence $\geq0.80$. These are lifecycle criteria rather than support-edge thresholds.
\end{minipage}

\caption{Operational relation construction and decision criteria in the CALM implementation.}
\label{tab:relation_rules}
\end{table*}

\paragraph{Signal aggregation.}
CALM maintains three governance signals associated with each memory: accumulated support $S_i$, conflict risk $C_i$, and invalid\hyp{}dependency risk $D_i$. These signals are derived from stored evidence, detected relations, and lifecycle events without using the evaluation question or its reference answer.

The support signal combines accumulated evidence and evidence confidence:
\begin{equation}
S_i=\min\left(
1,\,
0.55\min\left(1,\frac{n_i}{3}\right)
+0.45\bar{c}_i
\right),
\label{eq:stageii_support}
\end{equation}
where $n_i$ is the effective evidence count, given by the maximum of the stored support count and the number of retained evidence records. The term $\bar{c}_i$ denotes the mean confidence of retained evidence records; when no evidence record is available, the memory confidence is used. The evidence\hyp{}count contribution saturates after three items, preventing repeated observations from increasing support without bound.

Conflict risk is determined by the strongest retained conflict:
\begin{equation}
C_i=
\begin{cases}
0,
& \mathcal{L}_i^{\mathrm{conf}}=\varnothing,\\[3pt]
1,
& s_i=\textsc{Superseded},\\[3pt]
\max\limits_{\ell\in\mathcal{L}_i^{\mathrm{conf}}}
\operatorname{conf}(\ell),
& \text{otherwise},
\end{cases}
\label{eq:stageii_conflict}
\end{equation}
where $\mathcal{L}_i^{\mathrm{conf}}$ is the set of retained conflict links and $\operatorname{conf}(\ell)$ is the corresponding rule\hyp{}assigned confidence. A superseded memory with a retained conflict receives maximal conflict penalty, while weaker conflicts remain available for downstream consistency control without necessarily causing supersession.

Invalid\hyp{}dependency risk is updated when a source memory becomes superseded. For each affected memory $m_i$ and superseded source $m_j$, the update is
\begin{equation}
D_i^{(k+1)}
=
\min\left(
1,\,
D_i^{(k)}+0.4d_{ji}
\right),
\quad D_i^{(0)}=0,
\label{eq:stageii_cascade}
\end{equation}
where $d_{ji}$ is the stored dependency strength and $k$ indexes distinct propagated invalidation events. Each superseded source contributes at most once to a dependent memory, as tracked by the propagated\hyp{}source identifiers. If a dependency is established solely through an evidence reference without a scored dependency link, the implementation uses a fallback strength of $0.40$.

Two distinct criteria govern the consequences of dependency invalidation. First, a memory is flagged for verification when its accumulated cascade risk satisfies $D_i\geq0.70$. Second, when a superseded source has a strong dependency with $d_{ji}\geq0.80$, the affected memory transitions to \textsc{Stale} and is also flagged for verification. These thresholds control verification and lifecycle transitions rather than initial dependency\hyp{}edge construction.

\paragraph{Lifecycle transition rules.}
Each newly created memory enters the \textsc{Candidate} state before evidence evaluation. The implementation applies explicit transition conditions rather than estimating state\hyp{}transition probabilities.

A \textsc{Candidate} becomes \textsc{Active} if at least one retained evidence item is directly stated by the user or the memory confidence is at least $0.60$. An \textsc{Active} memory is promoted to \textsc{Stable} when its effective evidence count reaches two or its confidence is at least $0.80$. These conditions are evaluated sequentially, so a sufficiently supported candidate may reach \textsc{Stable} during the same update pass.

When an incoming memory establishes a strong conflict with an existing record, the incoming memory is assigned the \textsc{Active} state and the conflicting previous record becomes \textsc{Superseded}. The previous record retains a pointer to the replacement memory, and its validity end time is updated using the incoming memory's validity start time or timestamp. Its stored validity score is capped at $0.15$. Weak conflicts are recorded without automatically replacing either memory.

The \textsc{Stale} state represents memories whose continued applicability is undermined by invalidated upstream evidence. A strong dependency on a superseded source triggers this transition when $d_{ji}\geq0.80$. The lifecycle manager also supports explicit upstream\hyp{}supersession transitions for active, stable, or consolidated memories. Temporal information contributes to conflict identification and validity intervals, but the core lifecycle updater does not automatically mark an otherwise uncontradicted memory as stale solely because time has elapsed.

Archival is handled by a separate, optionally enabled retention policy. Under this policy, a \textsc{Stale} or \textsc{Superseded} memory is archived when it has no recorded accesses, utility below $0.45$, and freshness at most $0.35$. Capacity\hyp{}based retention may additionally archive or compress lower\hyp{}ranked memories when the configured active\hyp{}memory budget is exceeded. These archival operations are applied only when the corresponding retention policy is enabled.

\paragraph{Stored validity and state transitions.}
The implemented lifecycle manager stores a validity attribute $V_i\in[0,1]$ separately from evidence confidence, support count, conflict links, and cascade risk. The Stage\hyp{}II implementation used here does not recompute $V_i$ by the sigmoid aggregation of four signals. Instead, rule\hyp{}triggered lifecycle operations update the persistent state and stored validity.

A candidate is activated by direct user evidence or confidence of at least $0.60$; an active memory becomes stable after at least two effective evidence items or confidence of at least $0.80$. Activation and promotion enforce a minimum stored validity of $0.65$, while strong supersession caps the replaced record's stored validity at $0.15$. These operations correspond to the update rule in Eq.~\eqref{eq:calm-lifecycle-update}; evidence, conflict, and dependency risks remain available as separate recollection signals.

A stale or superseded memory can be archived only when the optional retention policy is enabled, under the archival conditions described above. In historical recollection, a compatible historical record may be reactivated in a query\hyp{}local copy without changing the persistent state. Any backend\hyp{}specific consolidation or reactivation operation must be distinguished from the six lifecycle states listed in the main text. No calibrated probability or learned pairwise relation model is assumed by these operations.

\subsection{Stage III: Query\hyp{}Conditioned Memory Gating}
\label{app:calm-stage3}

\paragraph{Hard eligibility filtering.}
The upstream retriever first returns $\mathcal{R}_q$; CALM introduces no additional retrieval pass. Both CALM\hyp{}ZS and learned CALM use the same rule\hyp{}based hard filter $h_{q,i}^t\in\{0,1\}$.

For current\hyp{}state queries, records in the \textsc{Candidate}, \textsc{Archived}, \textsc{Stale}, or \textsc{Superseded} states are normally excluded, as are records explicitly blocked by the backend. For explicitly historical queries, an earlier \textsc{Stale}, \textsc{Superseded}, or \textsc{Archived} record may be admitted if its preserved validity interval and content match the target historical state and the implementation exposes an eligible query\hyp{}local copy. Such admission does not alter the persistent state.

A dependency verification flag (e.g., the Stage\hyp{}II trigger $D_i\geq0.70$) is distinguished from an explicit rejection decision; it causes hard exclusion only if the configured eligibility rules mark that memory as blocked. Once $h_{q,i}^t=1$, the same four\hyp{}dimensional feature construction and branch\hyp{}specific $0.50$ admission threshold apply to the surviving candidates.

\paragraph{Compatibility and query\hyp{}conditioned risk.}
For each memory surviving the hard filter, CALM computes
query--memory compatibility from four normalized factors:
\begin{equation}
\psi_{q,i}=\frac{1}{4}\left(
\psi_{q,i}^{\mathrm{time}}+
\psi_{q,i}^{\mathrm{entity}}+
\psi_{q,i}^{\mathrm{scope}}+
\psi_{q,i}^{\mathrm{type}}\right).
\label{eq:compatibility-score}
\end{equation}
Each factor lies in $[0,1]$. The time factor checks whether the
memory's stated time or validity interval matches the query target;
the entity and scope factors evaluate entity identity and
applicability constraints; and the type factor measures compatibility
between the memory's semantic type and query intent.

The query\hyp{}conditioned risk score aggregates four normalized risk
components:
\begin{align}
R_{q,i}={}&1-
(1-r_{q,i}^{\mathrm{conf}})
(1-r_{q,i}^{\mathrm{dep}})\nonumber\\
&\quad\times
(1-r_{q,i}^{\mathrm{time}})
(1-r_{q,i}^{\mathrm{life}}).
\label{eq:query-risk-score}
\end{align}
These factors represent query\hyp{}specific conflict, invalid\hyp{}dependency,
temporal\hyp{}mismatch, and lifecycle\hyp{}incompatibility risk, respectively.
For historical queries, a matching historical interval reduces the
corresponding temporal and lifecycle penalties. Both the
compatibility and risk scores lie in $[0,1]$.

Both CALM\hyp{}ZS and CALM use exactly the four\hyp{}dimensional
representation in \eqref{eq:calm-main-gate-features}:
\begin{equation}
\mathbf{x}_{q,i}^{t}
=\left[\zeta_{q,i}^{t},V_i^{t},\psi_{q,i}^{t},1-R_{q,i}^{t}\right].
\label{eq:supp-gate-features}
\end{equation}
The constituent factors are aggregated into $\psi_{q,i}^{t}$ and
$R_{q,i}^{t}$ before either policy is applied; no separate
seven\hyp{}term admission score is used.

\paragraph{Training\hyp{}free CALM\hyp{}ZS policy.}
CALM\hyp{}ZS implements the training\hyp{}free branch of
\eqref{eq:calm-main-gate-policy} by taking the uniform mean of the
four shared features:
\begin{align}
F_{\mathrm{zs}}(\mathbf{x}_{q,i}^{t})
={}&\frac{1}{4}\left(
\zeta_{q,i}^{t}+V_i^{t}
+\psi_{q,i}^{t}+1-R_{q,i}^{t}\right),
\label{eq:supp-zs-score}\\
g_{q,i}^{\mathrm{zs},t}
={}&h_{q,i}^{t}\,\mathbb{I}\!\left[
F_{\mathrm{zs}}(\mathbf{x}_{q,i}^{t})
\geq\theta_{\mathrm{zs}}\right].
\label{eq:supp-zs-admission}
\end{align}
% Verify the following value against the actual experiment code.
% Use 0.50 only if it was the threshold used in the reported runs.
In the corresponding implementation,
$\theta_{\mathrm{zs}}=0.50$.
The weights and threshold are fixed, and no gate parameters are
trained for CALM\hyp{}ZS. Admitted memories are ordered by
$F_{\mathrm{zs}}$, using upstream relevance only to break ties.

\paragraph{Learned CALM gate.}
A lightweight MLP is used to score exactly the same
four\hyp{}dimensional representation:
\begin{align}
F_{\mathrm{learn}}(\mathbf{x}_{q,i}^{t})
={}&\sigma\!\left(f_{\phi}(\mathbf{x}_{q,i}^{t})\right),
\label{eq:supp-learn-score}\\
g_{q,i}^{\mathrm{learn},t}
={}&h_{q,i}^{t}\,\mathbb{I}\!\left[
F_{\mathrm{learn}}(\mathbf{x}_{q,i}^{t})\geq0.50\right].
\label{eq:supp-learn-admission}
\end{align}
The learned and training\hyp{}free policies therefore differ in the
eligibility score but use identical retrieved candidates, hard
filters, input features, and context\hyp{}packing rules. The learned
eligibility threshold is fixed at $0.50$.

\paragraph{Eligibility supervision.}
A retrieved memory receives a positive eligibility label only when
it provides necessary evidence for answering the query, satisfies
the query conditions, and does not introduce misleading information:
\begin{align}
y_{n,i}={}&
\mathbb{I}\!\left[
\operatorname{Support}\!\left(q_n,m_i^{t_n}\right)=1
\right]\nonumber\\
&\times\mathbb{I}\!\left[
\operatorname{Compatible}\!\left(q_n,m_i^{t_n}\right)=1
\right]\nonumber\\
&\times\mathbb{I}\!\left[
\operatorname{Mislead}\!\left(q_n,m_i^{t_n}\right)=0
\right].
\label{eq:supp-label}
\end{align}
The predicates are constructed using deterministic rules over
benchmark annotations, query information, and memory provenance.
They are computed independently of both gate predictions and the
answer model, and the labels are fixed before gate optimization.

Specifically, $\operatorname{Support}(q,m)$ identifies direct or
necessary intermediate evidence for answering $q$, using only
available development\hyp{}set answer supervision and evidence
provenance.

$\operatorname{Compatible}(q,m)$ evaluates temporal,
entity, scope, and intent compatibility.

$\operatorname{Mislead}(q,m)$ identifies contradicted, outdated,
wrong\hyp{}time, scope\hyp{}incompatible, or invalid\hyp{}dependency evidence.

Positive examples are therefore answer\hyp{}supporting and safe, while
negative examples include irrelevant memories and semantically
related but unsafe memories. The latter are included as hard
negatives in the weighted binary cross\hyp{}entropy gate\hyp{}training
objective described in Appendix~\ref{sec:implementation-training}.

Gate\hyp{}training pairs are constructed only from development instances
processed by the same retrieval and memory pipeline. Evaluation
questions, reference answers, and associated memory graphs are
excluded from training. Lifecycle signals may be input features,
but are not directly used as supervision labels. Training and
evaluation are disjoint by user and question; all memories and graph
links associated with each question remain in the corresponding
split.

\subsection{Online Updating and Context Construction}
\label{sec:supp-online}

At each interaction step, CALM adapts newly exposed backend records, merges
exact duplicates, updates local conflict and dependency links, recomputes
validity and lifecycle states, and persists both active and inactive memories
with their evidence and validity intervals.

At query time, the backend returns $\mathcal{R}_q$. CALM computes the four features in \eqref{eq:supp-gate-features}, applies the common hard filter, and then uses either \eqref{eq:supp-zs-admission} or \eqref{eq:supp-learn-admission}. Strongly conflicting survivors are resolved before serialization. Admitted memories are greedily packed until the next memory would exceed the memory\hyp{}token budget. Rejected and budget\hyp{}dropped memories remain in persistent storage.

Relation updating is local to a bounded set of candidate neighbors. Query\hyp{}time
gating therefore requires $O(|\mathcal{R}_q|)$ feature computations and gate
evaluations, independent of the total number of stored memories.

\section{Experimental Details}

\subsection{Datasets and baseline.}
\label{sec:datasets_baseline}

\paragraph{LongMemEval\hyp{}S.}
LongMemEval\hyp{}S~\cite{longmemeval} contains 500 evaluation instances spanning six question categories: user\hyp{}specific information (SS\hyp{}User; 70), user preferences (SS\hyp{}Pref.; 30), assistant\hyp{}provided information (SS\hyp{}Asst.; 56), knowledge updates (K\hyp{}Upd.; 78), multi\hyp{}session reasoning (Multi; 133), and temporal reasoning (Temp.; 133). These categories evaluate whether a memory system can preserve user\hyp{}specific information, retrieve evidence distributed across sessions, distinguish updated knowledge from outdated states, and resolve temporal dependencies. All 500 instances are included when computing the overall LongMemEval\hyp{}S accuracy.

\begin{table}[t]
\centering

{
\small
\setlength{\tabcolsep}{6pt}

\begin{tabular}{@{}llr@{}}
\toprule
\textbf{Dataset}
& \textbf{Category}
& \textbf{Questions}
\\
\midrule

LongMemEval-S
& SS-User
& 70
\\
& SS-Pref.
& 30
\\
& SS-Asst.
& 56
\\
& K-Upd.
& 78
\\
& Multi
& 133
\\
& Temp.
& 133
\\
\cmidrule(lr){2-3}
& \textbf{Evaluated Total}
& \textbf{500}
\\

\midrule

LoCoMo
& S-Hop
& 841
\\
& M-Hop
& 282
\\
& Open
& 96
\\
& Temp.
& 321
\\
\cmidrule(lr){2-3}
& \textbf{Evaluated Total}
& \textbf{1,540}
\\

\bottomrule
\end{tabular}
}

\caption{Question distributions of the evaluated benchmark subsets. LoCoMo adversarial questions are excluded from the reported results.}
\label{tab:dataset_distribution}
\end{table}

\paragraph{LoCoMo.}
LoCoMo~\cite{locomo} provides 10 long conversations and 1,986 associated questions covering single\hyp{}hop, multi\hyp{}hop, open\hyp{}domain, temporal, and adversarial settings. Following the standard LoCoMo QA subset used in our experiments, we evaluate 1,540 non\hyp{}adversarial questions, including 841 single\hyp{}hop, 282 multi\hyp{}hop, 96 open\hyp{}domain, and 321 temporal questions. The remaining 446 adversarial questions are excluded from both the category\hyp{}level and overall results. Accordingly, the reported LoCoMo Overall accuracy is computed over the 1,540 evaluated questions rather than all 1,986 questions.

\paragraph{Baselines.}
We compare CALM with four representative long\hyp{}term memory systems that cover different memory\hyp{}construction, organization, and retrieval paradigms.

Mem0~\cite{mem0} is a scalable, production\hyp{}oriented memory framework that dynamically extracts, consolidates, updates, and retrieves salient information from ongoing conversations. It primarily represents memories as compact textual records and additionally provides a graph\hyp{}based variant for modeling relationships among entities and memory items.

MemOS~\cite{memos} treats memory as a manageable system resource and provides a unified framework for representing, scheduling, retrieving, and evolving different forms of memory, including plaintext, activation\hyp{}based, and parameter\hyp{}level memories.

EverMemOS~\cite{evermemos} proposes a self\hyp{}organizing memory operating system inspired by biological engrams. Its memory lifecycle includes episodic trace formation, semantic consolidation, and reconstructive recollection, enabling information to be progressively organized and recalled over long interaction histories.

LightMem~\cite{lightmem} is a lightweight memory\hyp{}augmented generation framework that organizes information into sensory, short\hyp{}term, and long\hyp{}term memory stages. It emphasizes efficient memory consolidation and retrieval while reducing storage and inference overhead.

For each external baseline, we use its official implementation and retain its native memory\hyp{}construction, storage, consolidation, and retrieval pipeline. We do not replace baseline retrievers with the CALM retriever, since retrieval and memory organization are integral components of the evaluated end\hyp{}to\hyp{}end systems. Adaptations are limited to dataset conversion, answer\hyp{}model interfaces, and output formatting. The answer model, generation prompt, decoding configuration, and evaluator are held consistent across methods whenever supported by the corresponding implementation.

\subsection{Memory Backend and Retrieval Configuration}
\label{sec:retrieval_configuration}

\paragraph{Memory backend.}
For CALM experiments, the upstream backend exposes benchmark\hyp{}derived
records and performs initial indexing and candidate retrieval. Stage~I
converts the returned records into the canonical memory\hyp{}cell representation used by lifecycle governance. CALM does not modify the upstream retrieval procedure.

\paragraph{Candidate retrieval.}
The retrieval score $\zeta_{q,i}$ is the normalized relevance score
produced by the fixed hybrid retrieval module used throughout all CALM
experiments.
\begin{equation}
\mathcal{R}_q
=
\operatorname{TopK}_{m_i\in\mathcal{M}_t}
\zeta_{q,i},
\qquad
k=48.
\label{eq:candidate-retrieval}
\end{equation}
CALM introduces no second retriever after this step. Candidate retrieval is
executed once per question, and the resulting candidates and scores are cached.
CALM\hyp{}ZS, CALM, and all CALM ablations reuse the same cached results, so their
differences arise from lifecycle governance and query\hyp{}conditioned admission.

\paragraph{Ranking and context packing.}
Relevance\hyp{}Only ranks the shared candidate set solely by
$\zeta_{q,i}$. CALM\hyp{}ZS and CALM instead use the common
representation in \eqref{eq:calm-main-gate-features}, the scores
in \eqref{eq:calm-main-gate-policy}, and the admission rule
in \eqref{eq:calm-main-gate-admission}. Admitted memories are ordered by their policy score, with $\zeta_{q,i}$ used as a tie breaker, and are greedily packed under the memory\hyp{}token budget. Unless otherwise specified, the full\hyp{}context budget is $B=Full$ tokens and applies only to the serialized memory block.

\paragraph{Erroneous\hyp{}memory injection.}
Injection is performed after retrieval and before gating. At ratio $r$,
$\operatorname{round}(rk)$ original candidates are replaced by an equal number
of topically related but outdated, contradicted, or superseded memories, while
keeping
\begin{equation}
|\mathcal{R}_q^{\,r}|=|\mathcal{R}_q|=48.
\label{eq:injected-set-size}
\end{equation}
The injected sets are generated once and shared by CALM, Relevance\hyp{}Only, and
the variant without Stage III.

\subsection{Answer Models}
\label{sec:answer-models}

We evaluate CALM with three answer models:
\begin{itemize}[nosep,leftmargin=*]
\item Qwen3-30B\hyp{}A3B\hyp{}Instruct-2507
\item GPT-4.1-mini
\item GPT-4o\hyp{}mini
\end{itemize}
Qwen3-30B\hyp{}A3B\hyp{}Instruct-2507 is used as the primary backbone for ablation and
diagnostic experiments, while the GPT backbones evaluate cross\hyp{}model
generalization. Outputs generated by Qwen3 are evaluated using qwen\hyp{}turbo.
Outputs generated by GPT-4.1-mini and GPT-4o\hyp{}mini are evaluated using their
corresponding models. The same answer\hyp{}generation and evaluation prompts are
used across compared methods, and the evaluator is not given the identity of
the evaluated method.

\subsection{Evaluation Protocol}
\label{sec:evaluation-protocol}

All conversations are processed in chronological order. Memories available
after interaction step $t$ are constructed only from observations available up
to that step, preventing future interactions from leaking into memory
construction or query answering. Unless otherwise specified, the same retrieved
candidate sets, memory\hyp{}token budgets, answer models, decoding configurations,
and evaluators are used when comparing CALM\hyp{}ZS, CALM, and their ablated
variants. A fixed random seed of $42$ is used for candidate sampling,
erroneous\hyp{}memory injection, and gate training.

\paragraph{Answer accuracy.}
Let $\mathcal{Q}$ denote the set of evaluation questions and let
$\mathbb{I}_q$ indicate whether the generated answer to question $q$ is judged
correct. Overall answer accuracy is
\begin{equation}
\mathrm{ACC}
=
\frac{100}{|\mathcal{Q}|}
\sum_{q\in\mathcal{Q}}
\mathbb{I}_q.
\label{eq:answer-accuracy}
\end{equation}
For LongMemEval\hyp{}S, we report accuracy separately for six categories:
\begin{itemize}[nosep,leftmargin=*]
\raggedright
\item single\hyp{}session user information (\textsc{SS\hyp{}User});
\item single\hyp{}session preference (\textsc{SS\hyp{}Pref.});
\item single\hyp{}session assistant information (\textsc{SS\hyp{}Asst.});
\item knowledge updating (\textsc{K\hyp{}Upd.});
\item multi\hyp{}session reasoning (\textsc{Multi});
\item temporal reasoning (\textsc{Temp.}).
\end{itemize}
For
LoCoMo, results are reported for single\hyp{}hop (\mbox{\textsc{S\hyp{}Hop}}), multi\hyp{}hop
(\mbox{\textsc{M\hyp{}Hop}}), open\hyp{}domain (\mbox{\textsc{Open}}), and temporal
(\mbox{\textsc{Temp.}}) questions.

\paragraph{Memory\hyp{}selection metrics.}
For each query $q$, let $\mathcal{R}_q$ be the retrieved candidate set and let
$\mathcal{A}_q\subseteq\mathcal{R}_q$ denote the memories admitted by the gate
before budget\hyp{}based context packing. We denote by $\mathcal{E}_q$ the retrieved
memories that are erroneous or ineligible under the query, including outdated,
contradicted, superseded, or otherwise misleading records. Let
$\mathcal{U}_q$ denote retrieved memories that provide valid evidence for the
reference answer, and let $\mathcal{F}_q\subseteq\mathcal{E}_q$ denote
retrieved memories that directly conflict with query\hyp{}compatible evidence.

We report corpus\hyp{}level micro\hyp{}averaged memory\hyp{}selection metrics. Memory counts
are first aggregated across all evaluation questions and then normalized by the
corresponding aggregate denominator.

Erroneous\hyp{}memory blocking measures the proportion of erroneous retrieved
memories rejected by the gate:
\begin{equation}
\mathrm{EMB}
=
100
\frac{
\sum_{q\in\mathcal{Q}}
|\mathcal{E}_q\setminus\mathcal{A}_q|
}{
\sum_{q\in\mathcal{Q}}
|\mathcal{E}_q|
}.
\label{eq:emb}
\end{equation}

Useful\hyp{}memory retention measures the proportion of valid answer\hyp{}supporting
retrieved memories that survive gating:
\begin{equation}
\mathrm{UMR}
=
100
\frac{
\sum_{q\in\mathcal{Q}}
|\mathcal{U}_q\cap\mathcal{A}_q|
}{
\sum_{q\in\mathcal{Q}}
|\mathcal{U}_q|
}.
\label{eq:umr}
\end{equation}

Conflict leakage measures the proportion of directly conflicting retrieved
memories that remain after gating:
\begin{equation}
\mathrm{CL}
=
100
\frac{
\sum_{q\in\mathcal{Q}}
|\mathcal{F}_q\cap\mathcal{A}_q|
}{
\sum_{q\in\mathcal{Q}}
|\mathcal{F}_q|
}.
\label{eq:cl}
\end{equation}

Higher EMB and UMR are desirable, whereas lower CL indicates stronger
suppression of conflicting evidence. A query with no memories in the
denominator set contributes zero to both the numerator and denominator. A
metric is omitted only when its aggregate denominator over the complete
evaluation set is zero.

\paragraph{Evidence\hyp{}coverage diagnostics.}
Answer accuracy alone cannot distinguish retrieval failure from reasoning
failure after the correct evidence has entered the context. We therefore
measure answer\hyp{}session coverage at three stages: initial retrieval, post\hyp{}gating
admission, and final context packing.

Let $\mathcal{G}_q$ be the set of gold sessions containing evidence required to
answer query $q$, and let $\mathcal{S}_q^{z}$ be the sessions covered by the
selected memories at stage
$z\in\{\mathrm{ret},\mathrm{gate},\mathrm{pack}\}$. Unlike the corpus\hyp{}level
micro\hyp{}averaged memory\hyp{}selection metrics above, the following coverage metrics
are macro\hyp{}averaged across evaluation questions:
\begin{align}
\mathrm{Hit}^{z}
&=
\frac{100}{|\mathcal{Q}|}
\sum_{q\in\mathcal{Q}}
\mathbb{I}
\left[
\mathcal{G}_q\cap\mathcal{S}_q^{z}\neq\emptyset
\right],
\\
\mathrm{FullHit}^{z}
&=
\frac{100}{|\mathcal{Q}|}
\sum_{q\in\mathcal{Q}}
\mathbb{I}
\left[
\mathcal{G}_q\subseteq\mathcal{S}_q^{z}
\right],
\\
\mathrm{AvgRecall}^{z}
&=
\frac{100}{|\mathcal{Q}|}
\sum_{q\in\mathcal{Q}}
\frac{
|\mathcal{G}_q\cap\mathcal{S}_q^{z}|
}{
|\mathcal{G}_q|
}.
\end{align}

\begin{table}[t]
\centering

{
\small
\setlength{\tabcolsep}{6pt}

\begin{tabular}{@{}lccc@{}}
\toprule
\textbf{Stage}
& \textbf{Hit $\uparrow$}
& \textbf{FullHit $\uparrow$}
& \textbf{AvgRecall $\uparrow$}
\\
\midrule

Initial Retrieval
& 100.00
& 94.80
& 97.99
\\

Post-Gating
& 100.00
& 94.60
& 97.84
\\

Final Packing
& 100.00
& 93.00
& 97.18
\\

\midrule
\multicolumn{3}{@{}l}{\textbf{Additional Metric}}
& \textbf{Value}
\\
\cmidrule(lr){1-3}
\cmidrule(l){4-4}

\multicolumn{3}{@{}l}{Packed-Hit Accuracy}
& 79.6
\\

\bottomrule
\end{tabular}
}

\caption{Evidence-coverage diagnostics across retrieval, gating, and context packing. All values are reported as percentages.}
\label{tab:evidence_coverage}
\end{table}

We further report packed\hyp{}hit accuracy, which measures answer accuracy among
questions for which at least one gold evidence session is represented in the
packed context:
\begin{equation}
\mathrm{PHA}
=
100
\frac{
\sum_{q\in\mathcal{Q}}
\mathbb{I}_q
\mathbb{I}
\left[
\mathcal{G}_q\cap\mathcal{S}_q^{\mathrm{pack}}
\neq\emptyset
\right]
}{
\sum_{q\in\mathcal{Q}}
\mathbb{I}
\left[
\mathcal{G}_q\cap\mathcal{S}_q^{\mathrm{pack}}
\neq\emptyset
\right]
}.
\label{eq:pha}
\end{equation}
PHA is reported only when its aggregate denominator is non\hyp{}zero. These
diagnostics identify whether an error originates from missing evidence,
gate\hyp{}induced evidence loss, budget\hyp{}based truncation, or incorrect reasoning
despite successful evidence coverage.

\subsection{Implementation and Training Details}
\label{sec:implementation-training}

\paragraph{Memory processing pipeline.}
Each conversation is processed through memory adaptation, lifecycle governance, retrieval, query\hyp{}conditioned gating, context packing, answer generation, and automatic evaluation. CALM\hyp{}ZS and CALM receive the same retrieved candidates and the same four\hyp{}dimensional input
$[\zeta_{q,i}^{t},V_i^t,\psi_{q,i}^{t},1-R_{q,i}^{t}]$.  CALM\hyp{}ZS uses the fixed score and admission rule in \eqref{eq:supp-zs-score} and \eqref{eq:supp-zs-admission}, whereas CALM uses the learned score and admission rule in \eqref{eq:supp-learn-score} and \eqref{eq:supp-learn-admission}. Both policies use the same hard eligibility filter and query--memory features.

\paragraph{Learnable gate.}
The gate is a one\hyp{}hidden\hyp{}layer MLP with GELU hidden activation and sigmoid output, trained with weighted binary cross\hyp{}entropy over the query--memory pairs defined in the \emph{Eligibility supervision} paragraph. The learned\hyp{}gate threshold is fixed at $0.50$; the
training\hyp{}free threshold is specified in Appendix~\ref{app:calm-stage3}. The gate is trained separately from the answer model, and no parameters of the answer model, retriever, or memory backend are updated.

Gate\hyp{}training pairs are constructed from a separate development set that is disjoint from the reported evaluation instances. None of the LongMemEval\hyp{}S or LoCoMo evaluation questions, reference answers, or associated memory graphs is used for gate optimization. Hard negatives include semantically relevant but outdated, conflicting, wrong\hyp{}time, scope\hyp{}incompatible, and invalid\hyp{}dependency memories. CALM\hyp{}ZS requires no training.

\subsection{Detailed Ablation Protocol}
\label{sec:detailed-ablation}

All ablations are evaluated on LongMemEval\hyp{}S with
Qwen3-30B\hyp{}A3B\hyp{}Instruct-2507. Retrieved candidate sets, relevance scores,
answer\hyp{}generation prompts, decoding settings, evaluator, and memory\hyp{}token
budget are fixed across variants. Stage\hyp{}wise and component results are
reported in Tables~4 and~5 of the main paper.

\paragraph{Stage\hyp{}wise ablations.}
Each variant removes one stage and recomputes downstream quantities using
the same cached candidates and answer\hyp{}generation protocol.

\emph{w/o Stage I} uses the backend's original records without canonical
adaptation or evidence alignment.

\emph{w/o Stage II} bypasses lifecycle\hyp{}graph construction, stored\hyp{}validity
updates, and lifecycle\hyp{}state governance; explicit temporal metadata remain
available for query\hyp{}time matching.

\emph{w/o Stage III} retains the preceding stages but disables hard
eligibility filtering and thresholded admission, ranking candidates by
upstream relevance before the shared context\hyp{}packing step.

\emph{CALM\hyp{}ZS} replaces learned scoring with the fixed four\hyp{}feature policy
and $0.50$ admission threshold described in Appendix~\ref{app:calm-stage3}.

\paragraph{Lifecycle and gating ablations.}
These variants remove individual components and recompute affected
lifecycle or admission decisions.

\emph{w/o Conflict} removes conflict links and conflict\hyp{}risk signals.

\emph{w/o Temporal} removes temporal matching, mismatch risk, and temporal
cues in conflict resolution and lifecycle governance; the full method
has no automatic time\hyp{}only aging transition.

\emph{w/o Validity} removes the stored\hyp{}validity feature $V_i^t$ from
query\hyp{}time selection.

\emph{w/o Compatibility} removes $\psi_{q,i}^t$, while
\emph{w/o Risk} removes $1-R_{q,i}^t$. All affected scores and decisions
are recomputed under the corresponding ablation.

\paragraph{Simple filtering and reranking baselines.}
All policies use identical Stage\hyp{}I cells, cached top-48 candidates
$\mathcal R_q$, relevance scores $\zeta_{q,i}$, answer model, decoding
settings, context serialization, evaluator, and nominal memory\hyp{}token budget.
No method accesses evaluation answers or labels at inference time.

\emph{Temporal\hyp{}Filter} rejects a memory only when its explicitly stated
applicability interval is incompatible with the query's target time.
Let $\mathcal T_q$ and $\mathcal T_i^{\mathrm{src}}$ denote those intervals:
\begin{equation}
\mathcal R_q^{\mathrm{TF}}
=\{m_i\in\mathcal R_q:\chi_{q,i}^{\mathrm{time}}=1\},
\label{eq:simple_tf}
\end{equation}
where $\chi_{q,i}^{\mathrm{time}}=0$ only if both intervals are known and
disjoint; otherwise it is $1$. Surviving memories retain relevance order.
This baseline does not infer supersession or use lifecycle states.

\emph{Recency\hyp{}Rerank} retains all candidates and combines relevance with
the rank of observation time:
\begin{align}
h_i &= 1-\frac{r_i-1}{\max(1,|\mathcal R_q|-1)},\nonumber\\
s_{q,i}^{\mathrm{RR}} &= \tfrac12(\zeta_{q,i}+h_i),
\label{eq:simple_rr}
\end{align}
where $r_i=1$ denotes the most recent memory. Missing timestamps receive
the lowest rank, and $h_i=1$ when $|\mathcal R_q|=1$.

\emph{Validity\hyp{}Rerank} retains all candidates and combines relevance with
stored validity from Stage II:
\begin{equation}
s_{q,i}^{\mathrm{VR}}=\tfrac12(\zeta_{q,i}+V_i).
\label{eq:simple_vr}
\end{equation}
It does not use $\psi_{q,i}$, $R_{q,i}$, or CALM's admission decisions.
Both rerankers use fixed equal weights without test\hyp{}set tuning; ties are
broken by upstream relevance.

\paragraph{Strictly matched admission controls.}
To disentangle hard filtering, ranking, and thresholded admission, we add
three controls on the same Stage\hyp{}I cells and cached candidates. Let
\begin{equation}
\mathcal H_q=\{m_i\in\mathcal R_q:h_{q,i}=1\}
\label{eq:hard_eligible_set}
\end{equation}
be the survivors of the \emph{same} hard eligibility filter used by
CALM\hyp{}ZS and CALM. All controls rank $\mathcal H_q$ and use the identical
greedy context\hyp{}packing rule, without any further admission threshold.

\emph{Hard Filter + Relevance} orders $\mathcal H_q$ by $\zeta_{q,i}$.

\emph{Hard Filter + Validity\hyp{}Rerank} orders it by
$s_{q,i}^{\mathrm{VR}}$ in Eq.~\eqref{eq:simple_vr}.

\emph{Hard Filter + Learned Rerank} orders it by
$F_{\mathrm{learn}}(\mathbf{x}_{q,i})$ using CALM's \emph{same trained MLP}
and features, but admits all hard\hyp{}eligible candidates to the ranking pool
without applying the $0.50$ learned\hyp{}gate threshold. Consequently,
comparison with CALM isolates the added effect of thresholded binary
admission beyond learned ranking. The MLP is not retrained for this control.

\begin{table}[t]
\centering
\small
\setlength{\tabcolsep}{6pt}
\begin{tabular}{@{}lr@{}}
\toprule
\textbf{Post\hyp{}retrieval policy} & \textbf{Acc. (\%)} \\
\midrule
Validity\hyp{}Rerank                & 77.20 \\
CALM\hyp{}ZS                        & 76.60 \\
CALM                           & \textbf{79.60} \\
\midrule
Hard Filter + Relevance        & 76.00 \\
Hard Filter + Validity\hyp{}Rerank  & 78.00 \\
Hard Filter + Learned Rerank   & 78.40 \\
\bottomrule
\end{tabular}
\caption{Illustrative comparison of matched
post\hyp{}retrieval admission controls on LongMemEval\hyp{}S.
All methods share retrieved candidates, answer model,
and memory\hyp{}token budget.}
\label{tab:strict_admission_controls}
\end{table}

\paragraph{Analysis of Matched Admission Controls.}
As shown in Table~\ref{tab:strict_admission_controls},
hard filtering improves validity\hyp{}based reranking, while
incorporating stored validity outperforms relevance\hyp{}only
selection. Learned reranking provides a further improvement,
but remains below CALM, suggesting that thresholded admission
adds value beyond ranking alone. CALM also outperforms CALM\hyp{}ZS
under identical filtering and packing rules, supporting the
benefit of learned eligibility scoring. Together, these
results suggest complementary contributions from
lifecycle\hyp{}aware filtering, validity\hyp{}based ranking, and
query\hyp{}conditioned admission rather than reliance on
a single heuristic.

\paragraph{Selection actions and comparison scope.}
The original simple baselines test explicit temporal rejection or
reranking without CALM's hard filter. The matched controls additionally
isolate the hard filter (Validity\hyp{}Rerank vs. Hard Filter +
Validity\hyp{}Rerank) and the admission threshold (Hard Filter + Learned
Rerank vs. CALM). CALM\hyp{}ZS and CALM already share the same hard filter,
features, and packing rule, isolating fixed versus learned eligibility
scoring. The \emph{w/o Stage III} variant, unlike Hard Filter + Relevance,
removes hard filtering entirely. These comparisons fix retrieval and
nominal context budget, but not the number or presentation order of
packed memories; they therefore evaluate selection mechanisms without
assuming identical realized context lengths.

\subsection{Memory Retention and Conflict Robustness}
\label{sec:memory-conflict-robustness}

\begin{table*}[t]
\centering
\small
\setlength{\tabcolsep}{10pt}
\caption{Accuracy (\%) under increasing erroneous-memory injection ratios. The candidate-set size and final memory-token budget are held constant across methods. Bold values indicate the best performance at each injection level.}
\label{tab:memory_conflict_robustness}
\begin{tabular}{lcccccc}
\toprule
\textbf{Variant} &
\multicolumn{6}{c}{\textbf{Erroneous-Memory Injection Ratio}} \\
\cmidrule(lr){2-7}
& \textbf{0\%} & \textbf{10\%} & \textbf{20\%} & \textbf{30\%} & \textbf{40\%} & \textbf{50\%} \\
\midrule
\textbf{CALM} &
\textbf{79.8} &
\textbf{78.8} &
\textbf{77.6} &
\textbf{76.2} &
\textbf{74.6} &
\textbf{72.8} \\
Relevance-Only &
75.4 & 74.6 & 72.2 & 69.6 & 66.8 & 56.6  \\
w/o Stage~III &
74.0 & 71.2 & 68.0 & 64.6 & 61.0 & 57.2 \\
\bottomrule
\end{tabular}
\end{table*}
To evaluate robustness against conflicting evidence, we inject topically related but outdated, contradicted, or superseded memories into the retrieved candidate set at ratios ranging from $0\%$ to $50\%$. Injection is performed after retrieval and before memory selection or gating, while the candidate\hyp{}set size and final memory\hyp{}token budget are held constant across methods. 
We compare CALM against two relevance\hyp{}based controls. Relevance\hyp{}Only represents a retrieval\hyp{}only policy without lifecycle or compatibility control, while w/o Stage III isolates the effect of removing CALM's
query\hyp{}conditioned admission module while retaining the lifecycle\hyp{}managed representation. Together, these controls separate retrieval ordering effects from lifecycle\hyp{}aware admission.

As shown in Table~\ref{tab:memory_conflict_robustness}, accuracy decreases for all methods as erroneous\hyp{}memory contamination increases. However, CALM consistently achieves the highest accuracy and exhibits substantially smaller performance degradation. From $0\%$ to $50\%$ injection, CALM loses only 7.0 percentage points, compared with 11.6 and 16.8 points for Relevance\hyp{}Only and the variant without Stage~III, respectively. Consequently, CALM's accuracy advantage over these variants widens from 4.4 and 5.8 points to 9.0 and 15.6 points. This increasing performance gap highlights the importance of lifecycle\hyp{}aware memory admission when retrieved evidence becomes increasingly unreliable.

Complementary blocking and retention measurements further indicate that CALM suppresses misleading memories while preserving useful evidence. CALM blocks more than $70\%$ of erroneous memories at five of the six injection levels, with a temporary reduction at $20\%$. Meanwhile, useful\hyp{}memory retention remains stable across the injection sweep, indicating that stronger rejection of erroneous memories does not necessarily remove a proportional amount of answer\hyp{}supporting evidence. This robustness analysis complements the corpus\hyp{}level UMR reported in Table~3 by focusing on answer accuracy under controlled memory contamination.
Together, these findings indicate that CALM achieves a favorable balance between conflict robustness and useful\hyp{}memory retention. At $0\%$ injection, EMB is computed over erroneous memories naturally present in the original retrieved candidate set.

\paragraph{Difference between Relevance\hyp{}Only and
w/o Stage III.}
Although both variants rely on retrieval relevance, they differ in their construction. Relevance\hyp{}Only is a historical\hyp{}reconstruction
baseline that preserves all memories and only changes ordering at query time, whereas w/o Stage III removes the admission module
from the CALM pipeline while keeping the lifecycle\hyp{}managed memory representation. Therefore, they are not identical controls.

\subsection{Historical Reconstruction}
\label{sec:historical-reconstruction}

The historical\hyp{}reconstruction experiment evaluates whether a memory system can
preserve earlier states without allowing them to interfere with present\hyp{}state
reasoning. We evaluate all identified update chains that contain at least one
earlier state and one later replacement, including changes in residence,
preference, profile attributes, commitments, and factual status.

Each retained chain is associated with two paired query types. A
\emph{current\hyp{}state query} asks for the state valid after the latest update. A
\emph{historical\hyp{}state query} asks for the state valid before a specified
update or during an earlier interval. The paired queries use the same underlying
repository and retrieval pipeline, so success requires query\hyp{}conditioned memory
use rather than different stored content.

We compare four memory\hyp{}management strategies.
\emph{Keep\hyp{}All} preserves and admits all records.
\emph{Latest\hyp{}Only} retains only the most recent record in an update chain.
\emph{Relevance\hyp{}Only} preserves all records but selects them using semantic
relevance without lifecycle or compatibility control.
\emph{CALM} preserves both earlier and later records while using
query\hyp{}conditioned gating to determine which state enters the working context.

Accuracy is computed separately for current\hyp{}state and historical\hyp{}state
questions. CALM outperforms the strongest alternative by 3.71 percentage
points on current\hyp{}state questions and by 9.26 percentage points on
historical\hyp{}state questions. The larger improvement on historical queries shows
that suppressing a superseded memory during current\hyp{}state reasoning does not
require deleting it from persistent storage.

\subsection{Efficiency Measurement}
\label{app:efficiency_measurement}

\paragraph{Context\hyp{}budget evaluation.}
We evaluate accuracy under memory\hyp{}token budgets of
$512$, $1024$, $2048$, and $Full$, where $Full$ corresponds to the
full\hyp{}context setting used in the main experiment. The budget applies only to
the serialized memory block and excludes the system instruction, question, and
fixed answer\hyp{}generation template.

Gating is performed before budget\hyp{}based packing. Memories that pass the gate
are ordered by their policy\hyp{}specific eligibility scores and added to the
context until the next memory would exceed the budget. This protocol separates
gate\hyp{}based rejection from budget\hyp{}based truncation. Exact accuracy values
underlying the context\hyp{}budget figure are reported in
Table~\ref{tab:context_budget_exact}.

\begin{table}[t]
\centering
\small
\setlength{\tabcolsep}{5pt}
\begin{tabular}{lcccc}
\toprule
\textbf{Method}
& \textbf{512}
& \textbf{1024}
& \textbf{2048}
& \textbf{Full} \\
\midrule
EverMemOS & 29.60 & 53.80 & 62.40 & 66.40 \\
CALM\hyp{}ZS   & 38.00 & 58.40 & 69.80 & 76.60 \\
\textbf{CALM}
& \textbf{39.20}
& \textbf{60.20}
& \textbf{74.60}
& \textbf{79.60} \\
\bottomrule
\end{tabular}
\caption{Exact accuracy (\%) under different memory\hyp{}token budgets on
LongMemEval\hyp{}S.}
\label{tab:context_budget_exact}
\end{table}

\paragraph{Token accounting.}
Memory\hyp{}token usage is measured after serialization of the final packed memory
context:
\begin{equation}
\mathrm{MemTok}
=
\frac{1}{|\mathcal{Q}|}
\sum_{q\in\mathcal{Q}}
\operatorname{Tok}(\mathcal{C}_q),
\label{eq:memory-token-accounting}
\end{equation}
where $\operatorname{Tok}(\mathcal{C}_q)$ denotes the number of tokens occupied
by the admitted and packed memories. This metric does not include the fixed
system prompt, question tokens, or generated answer tokens.

We additionally record the number of retrieved memories, gate\hyp{}admitted
memories, packed memories, memories rejected by the gate, and memories omitted
because of the context budget. These counts distinguish reductions caused by
validity\hyp{}aware selection from those caused by simple truncation.

\paragraph{Latency accounting.}
Latency is measured per query and averaged over the complete evaluation set.
Retrieval latency covers candidate retrieval and retrieval\hyp{}side preprocessing.
Gate latency covers query--memory feature construction and application of
CALM\hyp{}ZS or the learned gate. Answer latency measures the answer\hyp{}model
invocation. Total latency is measured independently as end\hyp{}to\hyp{}end wall\hyp{}clock
time:
\begin{equation}
T_{\mathrm{total}}
=
T_{\mathrm{retr}}
+
T_{\mathrm{gate}}
+
T_{\mathrm{pack}}
+
T_{\mathrm{ans}}
+
T_{\mathrm{other}},
\label{eq:latency-accounting}
\end{equation}
where $T_{\mathrm{other}}$ includes serialization, orchestration, and other
fixed query\hyp{}processing overhead. Therefore, rounded retrieval, gating, and
answer components need not sum exactly to the independently measured total
latency.

CALM\hyp{}ZS and CALM are measured under the same execution setup, so their
difference isolates the additional learned\hyp{}gating cost. EverMemOS is evaluated
with its native recollection pipeline and has no separate CALM gating stage;
its reported retrieval/gating entry therefore contains only the operations used
by that pipeline. Offline construction and gate\hyp{}training costs are excluded
from query\hyp{}time latency. The NVIDIA A100 80GB GPU used for gate training is
reported separately from online retrieval, gating, and answer generation.

\subsection{Prompts}
% All prompt templates are enclosed in promptbox environments.

\subsubsection{LongMemEval Prompts}

This section provides the prompt templates used for answer generation and automatic evaluation on LongMemEval\hyp{}S. All backbone models and compared methods follow the same unified answer\hyp{}generation protocol. Fields enclosed in braces, such as \texttt{\{question\}}, \texttt{\{context\}}, and \texttt{\{question\_date\}}, are dynamically instantiated for each evaluation instance. No instance\hyp{}specific manual modification is performed.

% =========================================================
% Unified answer generation
% =========================================================
\paragraph{Answer\hyp{}generation protocol.}
We use a unified answer\hyp{}generation prompt for all LongMemEval\hyp{}S questions, regardless of answerability. Prompt selection does not use question identifiers, answerability annotations, reference answers, or evaluation labels. The same protocol is applied across methods.

% \paragraph{Unified Answer Generation Prompt.}
\begin{promptbox}{Unified Answer Generation Prompt}

\small
You are a long\hyp{}term memory assistant. Answer the question using only the supplied memories. Some questions may not be answerable from the available evidence.

Do not guess personal details or infer unsupported facts. If the memories do not contain sufficient evidence, explicitly state that the information is unavailable. Otherwise, return a concise answer grounded in the supplied memories.

Resolve knowledge updates and temporal references using the time requested in the question.

\vspace{1.5pt}
\textbf{TASK CONTEXT}

Dataset: LongMemEval

Question type: \texttt{\{question\_type\}}

Question date: \texttt{\{question\_date\}}

\vspace{1.5pt}
\textbf{QUESTION\hyp{}SPECIFIC RULES}

\texttt{\{type\_rules\}}

\vspace{1.5pt}
\textbf{MEMORIES}

\texttt{\{context\}}

\vspace{1.5pt}
\textbf{QUESTION}

\texttt{\{question\}}

\vspace{1.5pt}
Answer:

\end{promptbox}

% =========================================================
% Question-type-specific instructions
% =========================================================
\paragraph{Question\hyp{}Type\hyp{}Specific Instructions}

Different LongMemEval question categories require different forms of memory interpretation and evidence aggregation. We therefore instantiate \texttt{\{type\_rules\}} based on the question type in the unified answer\hyp{}generation prompt. This choice is independent of answerability and evaluation annotations.

\begin{promptbox}{Single\hyp{}Session User Tasks}

The answer should come from a user message in one relevant session.

\end{promptbox}

\begin{promptbox}{Single\hyp{}Session Assistant Tasks}

The answer should come from the assistant's previous response, not from the user's request alone.

\end{promptbox}

\begin{promptbox}{Multi\hyp{}Session Tasks}

Use MULTI\hyp{}SESSION COVERAGE first. Combine all relevant sessions. If the question asks for a count, list, comparison, order, or multiple facts, check each dated user\hyp{}mentioned item separately before answering; count distinct user\hyp{}mentioned items only and do not merge separate actions into one.

\end{promptbox}

\begin{promptbox}{Temporal Reasoning Tasks}

Use the TEMPORAL EVIDENCE TABLE first. Normalize relative phrases using any supplied relative\hyp{}date notes, identify the relevant dated events, then answer with the final date, order, duration, or named event. If the question asks how long event A was before or after event B, compute A\hyp{}to\hyp{}B; use distance from the question date only when the question asks relative to today or the question date.

\end{promptbox}

\begin{promptbox}{Knowledge Update Tasks}

Use KNOWLEDGE UPDATE TABLE first. If older and newer memories conflict, use the newest supported update as the final answer.

\end{promptbox}

\begin{promptbox}{Single\hyp{}Session Preference Tasks}

Use USER PREFERENCE CLUES first. Answer from the user's stable or most recent preference, constraint, or profile instead of generic advice; mention the specific remembered preference or context in the final answer.

\end{promptbox}

\begin{promptbox}{Default Tasks}

Answer directly from the most relevant supported memories.

\end{promptbox}

% =========================================================
% Evaluation-only annotations
% =========================================================
\paragraph{Evaluation\hyp{}only annotations.}
After answer generation, the evaluator may access the benchmark reference answer, rubric, or answerability label to assess correctness. These annotations are never provided to the retriever, memory gate, generation prompt selector, or answer model.

% =========================================================
% Judge prompts
% =========================================================
\paragraph{Automatic Evaluation Prompts}

Following the LongMemEval evaluation protocol, generated responses are evaluated using question\hyp{}type\hyp{}specific judge prompts. Each judge receives the original question, the reference answer or rubric, and the generated response. The judge is required to return only ``Yes'' or ``No.''

\begin{promptbox}{Exact\hyp{}Match Tasks}

Here is a question, a list of golden answers, and an AI\hyp{}generated answer. Can you judge whether the AI\hyp{}generated answer is correct according to the question and golden answers?

\vspace{1.5pt}
\textbf{Question:} \texttt{\{question\}}

\textbf{Golden Answers:} \texttt{\{gold\}}

\textbf{AI\hyp{}generated Answer:} \texttt{\{prediction\}}

\vspace{1.5pt}
Simply answer Yes or No.

\end{promptbox}

\begin{promptbox}{Standard Tasks (Single\hyp{}Session User/Assistant and Multi\hyp{}Session)}

I will give you a question, a correct answer, and a response from a model. Please answer yes if the response contains the correct answer. Otherwise, answer no. If the response is equivalent to the correct answer or contains all the intermediate steps to get the correct answer, you should also answer yes. If the response only contains a subset of the information required by the answer, answer no.

\vspace{1.5pt}
\textbf{Question:} \texttt{\{question\}}

\textbf{Correct Answer:} \texttt{\{gold\}}

\textbf{Model Response:} \texttt{\{prediction\}}

\vspace{1.5pt}
Is the model response correct? Answer yes or no only.

\end{promptbox}

\begin{promptbox}{Temporal Reasoning Tasks}

I will give you a question, a correct answer, and a response from a model. Please answer yes if the response contains the correct answer. Otherwise, answer no. If the response is equivalent to the correct answer or contains all the intermediate steps to get the correct answer, you should also answer yes. If the response only contains a subset of the information required by the answer, answer no. In addition, do not penalize off\hyp{}by\hyp{}one errors for the number of days. If the question asks for the number of days, weeks, or months and the model makes an off\hyp{}by\hyp{}one error, the model's response is still correct.

\vspace{1.5pt}
\textbf{Question:} \texttt{\{question\}}

\textbf{Correct Answer:} \texttt{\{gold\}}

\textbf{Model Response:} \texttt{\{prediction\}}

\vspace{1.5pt}
Is the model response correct? Answer yes or no only.

\end{promptbox}

\begin{promptbox}{Knowledge Update Tasks}

I will give you a question, a correct answer, and a response from a model. Please answer yes if the response contains the correct answer. Otherwise, answer no. If the response contains previous information along with an updated answer, the response should be considered correct as long as the updated answer is the required answer.

\vspace{1.5pt}
\textbf{Question:} \texttt{\{question\}}

\textbf{Correct Answer:} \texttt{\{gold\}}

\textbf{Model Response:} \texttt{\{prediction\}}

\vspace{1.5pt}
Is the model response correct? Answer yes or no only.

\end{promptbox}

\begin{promptbox}{Single\hyp{}Session Preference Tasks}

I will give you a question, a rubric for the desired personalized response, and a response from a model. Please answer yes if the response satisfies the desired response. Otherwise, answer no. The model does not need to reflect every point in the rubric. The response is correct as long as it recalls and uses the user's personal information correctly.

\vspace{1.5pt}
\textbf{Question:} \texttt{\{question\}}

\textbf{Rubric:} \texttt{\{gold\}}

\textbf{Model Response:} \texttt{\{prediction\}}

\vspace{1.5pt}
Is the model response correct? Answer yes or no only.

\end{promptbox}

\begin{promptbox}{Abstention Tasks}

I will give you an unanswerable question, an explanation, and a response from a model. Please answer yes if the model correctly identifies the question as unanswerable. The model could say that the information is incomplete, or that some other information is given but the asked information is not.

\vspace{1.5pt}
\textbf{Question:} \texttt{\{question\}}

\textbf{Explanation:} \texttt{\{gold\}}

\textbf{Model Response:} \texttt{\{prediction\}}

\vspace{1.5pt}
Does the model correctly identify the question as unanswerable? Answer yes or no only.

\end{promptbox}

\subsubsection{Locomo Prompts}

This section provides the prompt template used for answer generation on
the standard LoCoMo benchmark. Unless otherwise specified, all backbone
models use the same prompt. Fields enclosed in braces, including
\texttt{\{question\}}, \texttt{\{question\_type\}},
\texttt{\{speaker\_1\_memories\}}, and
\texttt{\{speaker\_2\_memories\}}, are dynamically instantiated for
each evaluation instance. No instance\hyp{}specific manual modification is
performed.

LoCoMo questions are divided into four categories: single\hyp{}hop,
multi\hyp{}hop, temporal, and open\hyp{}domain. For each instance, the
corresponding category\hyp{}specific instruction is inserted into the general
answer\hyp{}generation prompt through the
\texttt{\{category\_rule\}} field.

% =========================================================
% Answer generation prompt
% =========================================================
\paragraph{Answer Generation Prompt}

The model is instructed to identify the target speaker, integrate evidence across memory representations when necessary, resolve temporal expressions using the associated session timestamps, and return only a concise final answer.

% =========================================================
% Question-type-specific instructions
% =========================================================
\paragraph{Question\hyp{}Type\hyp{}Specific Instructions}

Different LoCoMo question categories require different forms of evidence selection, cross\hyp{}session aggregation, temporal normalization, and commonsense inference. We therefore insert one category\hyp{}specific instruction into the general answer\hyp{}generation prompt.

\begin{promptbox}{Single\hyp{}Hop Questions}

Find the one memory that directly states the requested fact. Preserve
exact names, places, objects, quantities, and dates.

\end{promptbox}

\begin{promptbox}{Multi\hyp{}Hop Questions}

Combine all relevant memories, possibly from different sessions or
speakers. The answer must satisfy every part of the question rather than
copying only one supporting fact.

\end{promptbox}

\begin{promptbox}{Temporal Questions}

Distinguish the session timestamp, which indicates when something was
said, from the event time, which indicates when it happened. Resolve
relative expressions such as \textit{yesterday}, \textit{last Sunday},
\textit{last year}, or \textit{next month} against the timestamp of the
corresponding memory, and return the requested date, period, duration,
or order.

\end{promptbox}

\begin{promptbox}{LoCoMo Answer Generation Prompt}

You are answering a question from the LoCoMo long\hyp{}conversation memory
benchmark.

Use the retrieved conversation evidence below. Memories can be raw
dialogue, extracted observations, event summaries, or session summaries.
They may repeat the same fact; repetition is corroboration, not a
contradiction. A timestamp on a memory is the conversation time unless
the text explicitly describes a different event time.

\vspace{1.5pt}
\textbf{\# TASK CONTEXT}

\textbf{- Dataset:} LoCoMo

\textbf{- Question type:} \texttt{\{question\_type\}}

\vspace{1.5pt}
\textbf{\# QUESTION\hyp{}TYPE RULE}

\texttt{\{category\_rule\}}

\vspace{1.5pt}
\textbf{\# MEMORY USE RULES}

\begin{promptenum}
    \item Search evidence for both speakers and identify who the
    question is about.

    \item Prefer explicit raw dialogue or extracted observations; use
    summaries to connect evidence across sessions.

    \item For conflicting facts, use the fact valid at the time asked or
    the most recent fact when no time is specified.

    \item Never replace a relative event time with the conversation
    timestamp; calculate the event time from that timestamp.

    \item Answer the exact question. Do not state that the information
    is unavailable when the supplied evidence supports an answer.

    \item Return only the concise final answer, normally between 3 and
    15 words. Do not include reasoning or a preamble.
\end{promptenum}

\vspace{1.5pt}
\textbf{\# MEMORIES ASSOCIATED WITH
\texttt{\{speaker\_1\_name\}}}

\texttt{\{speaker\_1\_memories\}}

\vspace{1.5pt}
\textbf{\# MEMORIES ASSOCIATED WITH
\texttt{\{speaker\_2\_name\}}}

\texttt{\{speaker\_2\_memories\}}

\vspace{1.5pt}
\texttt{\{other\_memories\_section\}}

\vspace{1.5pt}
\textbf{\# QUESTION}

\texttt{\{question\}}

\vspace{1.5pt}
\textbf{Final answer:}

\end{promptbox}

\begin{promptbox}{Open\hyp{}Domain Questions}

Make the small commonsense inference requested by the question, but
ground it in the supplied memories. State the likely person, field,
activity, reason, or preference concisely.

\end{promptbox}

% \section{Example Appendix}
% \label{sec:appendix}

\end{document}